\documentclass[lettersize,journal]{IEEEtran}
\usepackage{amsmath,amsfonts}
\usepackage{algorithmic}
\usepackage{algorithm}
\usepackage{array}
\usepackage[caption=false,font=normalsize,labelfont=sf,textfont=sf]{subfig}
\usepackage{textcomp}
\usepackage{stfloats}
\usepackage{url}
\usepackage{verbatim}
\usepackage{graphicx}
\usepackage{cite}
\usepackage{bbm}
\usepackage[utf8]{inputenc} 
\usepackage[T1]{fontenc}    
\usepackage[colorlinks,
            linkcolor=teal,      
            anchorcolor=teal, 
            citecolor=teal,       
            ]{hyperref}
\usepackage{url}            
\usepackage{booktabs}       
\usepackage{amsfonts}       
\usepackage{nicefrac}       
\usepackage{microtype}      
\usepackage{xcolor}         
\usepackage{xspace}         
\usepackage{wrapfig}     
\usepackage{makecell}
\usepackage{algorithm}
\usepackage{adjustbox}
\usepackage{colortbl}
\definecolor{Gray}{gray}{0.92}
\usepackage{amsmath}
\usepackage{amssymb}
\usepackage{mathtools}
\usepackage{amsthm}
\usepackage{multirow}
\usepackage{array} 
\usepackage{amssymb}
\usepackage{bbding}
\usepackage{pifont}
\usepackage{wasysym}
\usepackage{amssymb}
\usepackage{wrapfig}
\usepackage{enumitem}
\usepackage{makecell}
\usepackage{adjustbox}
\usepackage{colortbl}
\definecolor{Gray}{gray}{0.85}
\definecolor{sclgreyblue}{rgb}{0.2,0.3,0.5}%
\usepackage{amsmath}
\usepackage{amssymb}
\usepackage{mathtools}
\usepackage{amsthm}
\usepackage{multirow}
\usepackage{wrapfig}
\usepackage{enumitem}

\newcommand{\lightgrey}[1]{\textit{\color{sclgreyblue} #1}}

\newcommand*{\circled}[1]{\lower.7ex\hbox{\tikz\draw (0pt, 0pt)%
    circle (.5em) node {\makebox[1em][c]{\small #1}};}}

\newcommand{\abbr}[0]{AdaODD\xspace}

\newcommand{\Gray}[0]{\rowcolor{gray!20}}
 
\begin{document}

\title{Model-free Test Time Adaptation for Out-Of-Distribution Detection}

\author{YiFan Zhang, Xue Wang, Tian Zhou, Kun Yuan, \\ Zhang Zhang, Liang Wang~\IEEEmembership{Fellow,~IEEE}, Rong Jin
\thanks{Yi-Fan Zhang, Zhang Zhang, and Liang Wang are with State Key Laboratory of Multimodal Artificial Intelligence Systems (MAIS), Center for Research on Intelligent Perception and Computing (CRIPAC), Institute of Automation, Chinese Academy of Sciences (CASIA), Beijing 100190, China, and also with the School of Artificial Intelligence, University of Chinese Academy of Sciences (UCAS), Beijing 100049, China (e-mail: yifanzhang.cs@gmail.com).}
\thanks{Xue Wang, Tian Zhou are with Alibaba Group.}
\thanks{Kun Yuan is with Center for Machine Learning Research, Peking University.}
\thanks{Rong jin is affiliated with Meta, and work done at Alibaba Group.}
}

\markboth{Journal of \LaTeX\ Class Files,~Vol.~14, No.~8, August~2021}%
{Shell \MakeLowercase{\textit{et al.}}: A Sample Article Using IEEEtran.cls for IEEE Journals}


\maketitle

\begin{abstract}
    Out-of-distribution (OOD) detection is essential for the reliability of ML models. Most existing methods for OOD detection learn a fixed decision criterion from a given in-distribution dataset and apply it universally to decide if a data point is OOD. Recent work~\cite{fang2022is} shows that given only in-distribution data, it is impossible to reliably detect OOD data without extra assumptions. Motivated by the theoretical result and recent exploration of test-time adaptation methods, we propose a Non-Parametric Test Time \textbf{Ada}ptation framework for \textbf{O}ut-Of-\textbf{D}istribution \textbf{D}etection (\abbr). Unlike conventional methods, \abbr utilizes online test samples for model adaptation during testing, enhancing adaptability to changing data distributions. The framework incorporates detected OOD instances into decision-making, reducing false positive rates, particularly when ID and OOD distributions overlap significantly. We demonstrate the effectiveness of \abbr through comprehensive experiments on multiple OOD detection benchmarks, extensive empirical studies show that \abbr significantly improves the performance of OOD detection over state-of-the-art methods. Specifically, \abbr reduces the false positive rate (FPR95) by $23.23\%$ on the CIFAR-10 benchmarks and $38\%$ on the ImageNet-1k benchmarks compared to the advanced methods. Lastly, we theoretically verify the effectiveness of \abbr.




\end{abstract}
\section{Introduction}\label{sec:intro}
Traditional machine learning models often exhibit suboptimal performance when the training and test data come from different distributions. This weakness impedes the real-world deployment of machine learning systems, particularly in safety-critical applications such as autonomous driving~\cite{ramanagopal2018failing}, and biometric authentication~\cite{masi2018deep}. To mitigate the risk of out-of-distribution (OOD) data, the OOD detection problem has been studied \cite{hendrycks2016baseline,lee2018simple,yang2021generalized}, which requires an outlier detector to determine whether the input is ID (in-distribution) or OOD. For ID inputs, the system will predict their true class and  OOD inputs will be rejected and not be predicted. Various methods have been developed to improve the performance of OOD detection, including using the classification confidence or entropy~\cite{hendrycks2016baseline,liu2020energy}, modeling the ID density~\cite{zong2018deep,abati2019latent}, computing the feature distances~\cite{lee2018simple,wang2022vim}, and exposing to OOD samples while training~\cite{hendrycks2018deep}. 

Despite the encouraging progress, the performance achieved for OOD detection remains limited, especially for large-scale datasets. For example, for {large-scale ImageNet task~\cite{huang2021mos}}, which has $1,280,000$ labeled images, the false positive rate of OOD detection is only $53.97\%$ even using the state-of-the-art OOD detection method, namely, more than half of OOD samples have not been successfully detected. The challenge of OOD detection is further highlighted by the impossibility theorem recently developed for OOD detection~\cite{fang2022is}, i.e. it is impossible to reliably detect OOD data with access to only ID data points without making strong assumptions. 

\begin{figure}[t]
    \centering
    \minipage{0.48\linewidth}
  \includegraphics[width=\linewidth]{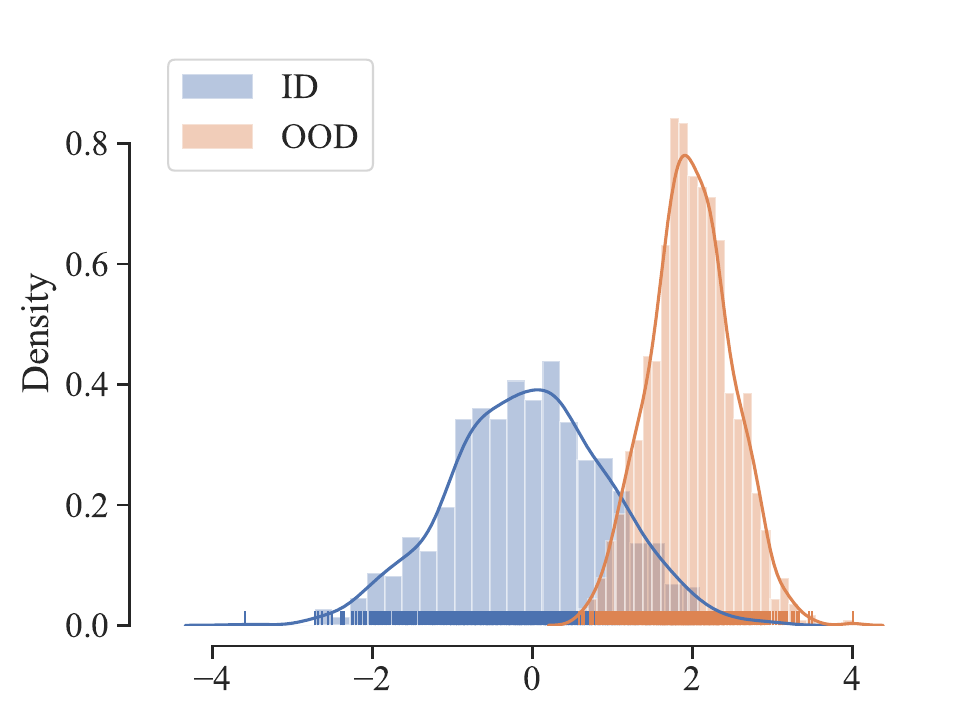}
\endminipage
\minipage{0.48\linewidth}%
  \includegraphics[width=\linewidth]{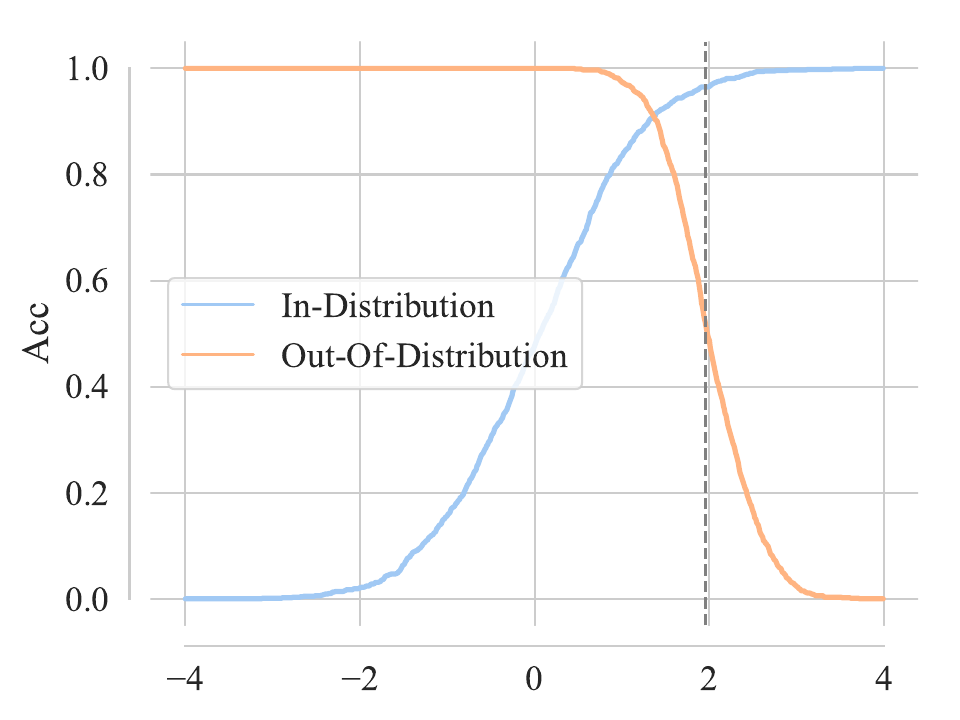}
\endminipage
\caption{\textbf{The motivation of using test-time adaptation}, (a) $\mathcal{D}^{in}=\mathcal{N}(0,1)$ and  $\mathcal{D}^{out}=\mathcal{N}(2,1/2)$. (b) shows the detection accuracy of both ID and OOD samples, considering different decision boundaries. The decision boundary $x=1.96$ that is learned from only ID data is depicted by the grey dash line.}
   \label{fig:motivation}\label{fig:motivation2}\label{fig:motivation2_acc}
\end{figure}

As indicated by the impossibility theorem from~\cite{fang2022is}, the fundamental challenge of OOD detection arises from the fact that no OOD data point is available. 
As a result, a fixed decision criterion is learned from the ID dataset and applied universally regardless of the OOD distribution. To further highlight this limitation, we construct an illustrative example in~\figurename~\ref{fig:motivation} where ID data points are sampled from a normal distribution $\mathcal{N}(0,1)$. Following typical OOD detection methods (e.g.,\cite{sun2022out}), we set the threshold to separate ID/OOD samples as 1.96 since it reaches $95\%$ confidence for the ID distribution.
It works well when the OOD distribution is far from the ID distribution. But when both distributions overlap significantly, as shown in ~\figurename~\ref{fig:motivation2}, this approach yields a much higher false positive rate (near $50\%$) for OOD samples, implying the necessity to adjust decision criteria based on test samples. One possible solution is to dynamically adjust the threshold based on the test examples that have been detected as OOD samples. 
In particular, for the example illustrated in Figure~\ref{fig:motivation}, using the detected OOD samples, we are able to decrease the value of the threshold, leading to a significant reduction in the false positive rate of OOD detection.\footnote{The purpose of \figurename~\ref{fig:motivation2_acc} is to demonstrate that if we have some initial estimations of the OOD data, we can adjust the learned decision boundary accordingly to achieve a better trade-off in the classification of ID and OOD samples. By utilizing the test-time adaptation approach, we can leverage online test data to improve the OOD detector's performance and adjust the decision boundary dynamically based on the changing data distribution during deployment. This adaptability allows the detector to better distinguish between ID and OOD samples and achieve improved overall performance. We acknowledge that the example with two Gaussians is relatively simple and may not fully capture the complexities of real-world scenarios. Nevertheless, it serves as a conceptual demonstration of the advantages of utilizing online test instances to improve OOD detection. For real-world utility, the experimental section provides superior detection results on many large-scale benchmarks.} (The detailed experimental setting is depicted in Section~\ref{sec:data_detail})
  
Motivated by the example, we propose to study OOD detection in a fully test-time adaptation setting~\cite{wang2020tent}, which assumes that test samples come from online data streams and can be used to adjust the model. 
Different from conventional studies on OOD detection, where the detected OOD samples cannot be fully utilized. Test-time adaptation (TTA) allows us to use online test samples, which more closely match realistic settings where a pre-trained model is supposed to be adapted to unlabeled data during testing, before making predictions. 
Intuitively, by using test samples labeled as OOD, we can approximately estimate the OOD distribution and adjust the decision criterion to improve the accuracy of OOD detection. However, existing TTA methods are mostly designed for classification tasks and use the pseudo-label to retrain the model in an online manner~\cite{zhang2022domain,wang2022continual,wang2020tent,zhang2023adanpc}. It is nontrivial to apply them for the OOD detection task, where the model is trained by only ID instances and the ID/OOD label spaces are disjoint. To this end, we propose a simple yet effective method, a non-parametric Test Time Adaptation framework for Out-Of-Distribution Detection, or \abbr for short, to best leverage online test samples that are pseudo-labeled by the existing decision criterion. Specifically, we construct a feature memory bank to include all feature vectors from training. For a test sample, we compute its score based on its $k$-nearest neighbors from the memory and label it as OOD when the score exceeds a given threshold. Any test sample labeled as OOD is included in the memory bank for future score calculation. To sum up, we make three contributions:

1. We investigate a non-parametric paradigm for performing TTA by storing feature maps of test instances. The proposed \abbr can be incorporated with any back-end model for OOD detection. We show that incorporating online detected OOD instances into decision-making can reduce the false positive rate, especially when In/Out-of-distribution has large overlapping. We justify the design of \abbr from the viewpoint of a maximum likelihood estimator. 

2. Comprehensive experiments on multiple OOD detection benchmarks are conducted. Our results show that \abbr can substantially reduce the false positive rate compared to state-of-the-art baselines, for example, \abbr reduces the FPR $38\%$ on ImageNet-1k benchmarks. 

3. We conduct thorough ablation studies on each component of \abbr. The results shed light on how we choose hyper-parameters considering different scales of datasets.

\section{Related Work}

\label{sec:related}

\textbf{Out-of-distribution detection methods} mainly include (1)  classification-based methods that derive OOD scores based on the classifier trained on ID data~\cite{hendrycks2016baseline}, such as using the maximum softmax probability~\cite{hendrycks2016baseline}, and energy~\cite{liu2020energy} as the OOD score. (2) Density-based methods model the in-distribution with some probabilistic models and data samples lie in low density regions are classified as OOD data~\cite{zong2018deep,abati2019latent,xiao2020likelihood}. and (3) Distance-based methods are based on an assumption that OOD samples should be relatively far away from representations of in-distribution classes~\cite{lee2018simple,techapanurak2020hyperparameter,wang2022vim}. 
Different from them, to our best knowledge, the proposed \abbr first seeks to utilize the online target data. Although there have been some out-of-distribution detection methods that utilize auxiliary OOD data. This line of work regularizes the model during training~\cite{katz2022training,liu2020energy,ming2022poem,du2022unknown,katz2022training}, which is different from our fully test-time adaptation setting, where we can exploit the benefit from the online test instances and attain much better performance. 

\textbf{Test-time adaptation methods} are recently proposed to utilize target samples. TTA methods can be divided into the following categories: Test-time training methods design proxy tasks during tests such as self-consistence~\cite{zhang2021test}, rotation prediction~\cite{sun2020test}, entropy minimization~\cite{wang2020tent,zhang2021memo} or update a prototype for each class~\cite{iwasawa2021test}; Domain adaptive methods~\cite{dubey2021adaptive} need additional models to adapt to target domains. Single sample generalization methods~\cite{xiao2022learning,zhang2022domain} need to learn an adaption strategy from source domains; the aforementioned approaches can only work when the label space of training and test data is the same. However, OOD detection problem has only ID data during training, and existing TTA methods cannot adapt to such a setting directly. We first propose \abbr that performs Test-time adaptation for OOD detection by storing the test samples. 

\textbf{Non-parametric TTA}. The most similar work to us is AdaNPC~\cite{zhang2023adanpc}, which uses a nonparametric k-nearest neighbors classifier for test-time adaptation. The differences between our paper and AdaNPC are as follows: a. \textbf{Focus and Task}: AdaNPC focuses on domain generalization in image classification, where the label space between training and inference is the same. In contrast, our paper deals with the OOD detection problem, where the training data contains only ID data, making the task inherently different from DG. Therefore, AdaNPC cannot be easily extended to OOD detection. b. \textbf{Design Mechanisms}: The main focus of our paper lies in designing mechanisms such as memory augmentation strategies and score functions to better distinguish features from OOD and ID. These design aspects are specific to OOD detection and aim to improve the overall performance of the detector. AdaNPC utilizes a naive KNN classifier and a simple first-in-first-out (FIFO) strategy for memory augmentation.

\textbf{Transfer learning theory.} The first line of work that considers bounding the error on target domains by the source domain classification error and a divergence measure, such as $d_\mathcal{A}$ divergence~\cite{ben2006analysis,david2010impossibility} and $\mathcal{Y}$ divergence~\cite{mansour2009domain,mohri2012new}. However, the symmetric differences carry the wrong intuition and most of these bounds depend on the choice of hypothesis~\cite{kpotufe2018marginal,zhang2022domain}. There are also some studies that consider the density ratio between the source and target domain~\cite{quinonero2008dataset,sugiyama2012density,zhang2022domain}, and transfer-exponent for non-parametric transfer learning~\cite{kpotufe2018marginal,cai2021transfer,hanneke2019value,reeve2021adaptive}.

\textbf{Semi-supervised learning.} This work also refers to semi-supervised learning methods, which use both labeled and unlabeled data for model training~\cite{yang2022survey}. \abbr assigns pseudo labels for online test samples and select reliable OOD samples for adaptation. The high level idea of pseudo-labeling methods~\cite{zhou2010semi,blum1998combining,qiao2018deep,dong2018tri} is similar to us. Differently, the unlabeled data in this paper is from an online data stream but not the offline dataset. Therefore, we cannot retrain the full model because of the computation requirement of TTA methods~\cite{wang2020tent} and the need for more lightweight operations.

\section{Methods}\label{sec:method}

\textbf{Notation.} Given an input space $\mathcal{X}$, representation space $\mathcal{Z}$, and label space $\mathcal{Y}=\{1,2,...,c\}$, we have an ID dataset ${D}_{in}=\{(x_i,y_i)\}_{i=1}^N$ that is drawn from distribution $\mathcal{D}^{in}$. We have an OOD distribution $\mathcal{D}^{out}$ that is different from $\mathcal{D}^{in}$. We also have a feature encoder $h_\theta:\mathcal{X}\rightarrow \mathcal{Z}$ and a classifier $f:\mathcal{Z}\rightarrow \mathbb{R}^c$, where $h_\theta,f$ are often parameterized by deep neural networks.

\textbf{Test-time Out-of-distribution (OOD) detection.} Given data $D_{in}$ sampled from ID $\mathcal{D}^{in}$, we train a neural network $h_\theta\circ f$. Our target is to construct a score function $S(x)$, which assigns higher scores for ID samples and lower scores for OOD samples. Formally, given a threshold $\lambda$ and input $x$, the decision that $x$ is from $\mathcal{D}^{in}$ is made by an indicator function $\mathbf{I}\{S(x)\geq \lambda\}$, otherwise, $x$ is classified as an OOD sample. In this paper, like previous studies of OOD detection (e.g. ~\cite{yang2021generalized}), we assume that the labels of samples from $\mathcal{D}^{out}$ do not overlap with $\mathcal{D}^{in}$, which is often referred to as label shift. Unlike existing studies~\cite{yang2021generalized}, during inference, we propose not only to detect the OOD instances but also to utilize these examples to improve the detection capability.

\subsection{Motivation of test-time adaptation.} 

\begin{figure}
    \centering
     \includegraphics[width=0.4\textwidth]{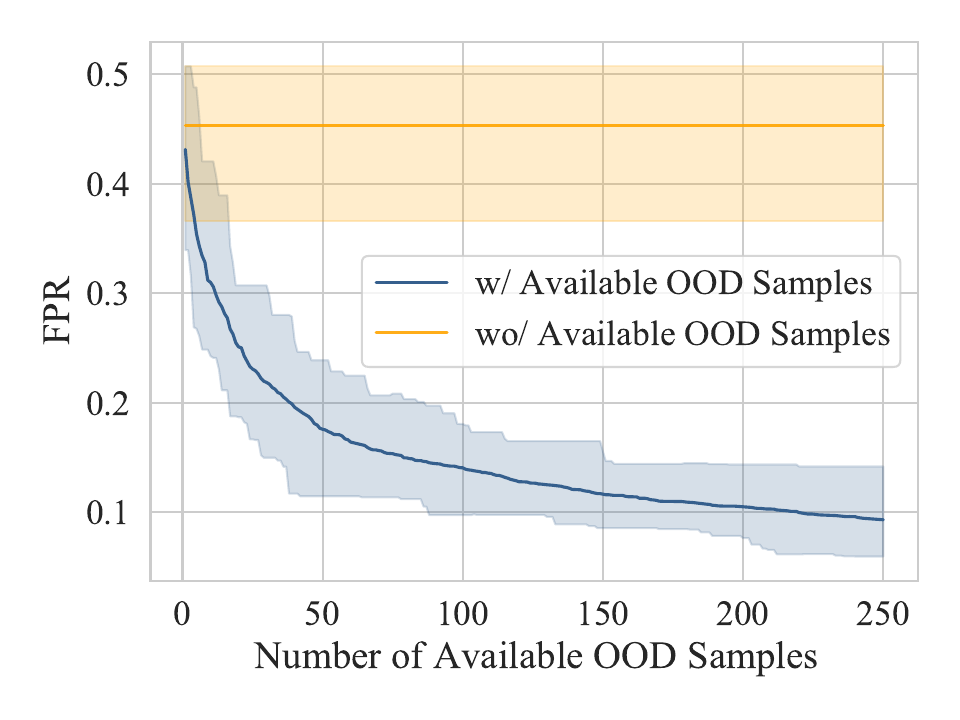}
    \caption{FPR considering different available OOD samples.}
        \label{fig:motivation_ood}
\end{figure}
As discussed in the introduction, without $\mathcal{D}^{out}$, a fixed decision threshold (i.e. $x = 1.96$) independent of test samples may not work well, particularly when the ID and OOD distributions overlap significantly. On the other hand, using the test samples labeled as OOD by our detection algorithm, we can obtain an estimation of $\mathcal{D}^{out}$ and reduce the decision threshold to $x=1.35$. It will result in the correct classification of $91\%$ ID samples and $90\%$ OOD samples, a significant improvement in OOD detection ($40\%$) at a small sacrifice of classification accuracy for ID samples ($4\%$).

To further quantitatively investigate the impact of OOD samples on the performance of OOD detection, in~\figurename~\ref{fig:motivation_ood}, we include the performance of OOD detection with various numbers of OOD samples from $\mathcal{D}^{out}=\mathcal{N}(2,1/2)$ (See Appendix~\ref{sec:data_detail} for the detailed experimental setting). 
The result shows that even with only a few OOD samples, our algorithm can dramatically reduce the FPR when compared to the detection algorithms that solely depend on ID samples.


\begin{algorithm*}[tb]
   \caption{Non-Parametric Test Time Adaptation for Out-Of-Distribution Detection}
   \label{alg:main}
\begin{algorithmic}[1]
   \STATE {\bfseries Input:} ID dataset ${D}^{in}$, a pretrained encoder $h_\theta$ with parameter $\theta$. Test sample $x_{o}$, threshold $\lambda$, $k$ for KNN selection. OOD selection margin $\gamma$ and  sample scale $\kappa$.
   \STATE {\bfseries Initialize} the memory bank with scales $\mathcal{M}=\{(z_i:=h_\theta(x_i)/\parallel h_\theta(x_{i}) \parallel_2, s_i:=1)\}_{x_i\in {D}^{in}}$.
   \STATE {\color{brown}\bfseries Inference stage for $x_{o}$.}
   
    \STATE  $z_{o}=h_\theta(x_{o})/\parallel h_\theta(x_{o}) \parallel_2$. \textit{ $\quad\quad\quad\quad\quad\;$// Normalize the feature of the test sample.} \\
    \STATE $B=\left((z_{(1)},s_{(1)}),...,(z_{(k)},s_{(k)})\right)$. \textit{$\quad\quad\quad$// Find the nearest neighbors of the test feature.}  \\
    \STATE $S(x_o)=\frac{1}{k}\sum_{i=1}^k -\parallel z_o-z_{(i)} \parallel s_{(i)}$ \textit{$\quad$// Calculate the score function.}
    \STATE  $\mathbf{1}\{S(x_o)\geq \lambda\}$ \textit{$\quad\quad\quad\quad\quad\quad\quad\quad\quad\quad\;$// Make OOD detection decision.} \label{alg1:decision}
   \STATE {\color{brown}\bfseries Memory augmentation for $x_{o}$.} 
   \IF{$S(x_o)< \lambda \gamma$ } 
   \STATE $\mathcal{M}=\mathcal{M}\bigcup \{(z_o, \kappa)\}$ \textit{ $\quad\quad\quad\quad\quad\;$// The test sample is more likely to be an OOD instance.}
   \ENDIF
   \IF{$S(x_o)> \lambda / \gamma$}
   \STATE $\mathcal{M}=\mathcal{M}\bigcup \{(z_o, 1)\}$ \textit{ $\quad\quad\quad\quad\quad\;$// The test sample is more likely to be an ID instance.}
   \ENDIF
\end{algorithmic}
\end{algorithm*}




\subsection{Non-Parametric Test Time Adaptation for Out-Of-Distribution Detection}

Here we describe \abbr and discuss its key merits compared to existing OOD detectors.


K-Nearest Neighbors (KNN)~\cite{sun2022out} is a distance-based OOD detection method, which utilizes feature embedding to compute distances and detect OOD samples. Compared to parametric density estimation methods, KNN is advantageous in that it does not impose strong distributional assumptions, and is efficient for large databases with billions of images. The proposed \abbr improves KNN~\cite{sun2022out} by \textbf{explicitly exploiting online test samples that are detected as OOD}. The full detail of \abbr is given in Algorithm~\ref{alg:main}. For a given test sample $x_o$, we first find its $k$ nearest neighbors in the memory bank $\mathcal{M}$, which is initialized from the normalized training features, and the scale $s_i$ of all ID samples are set to $1$. We then calculate the score $S(x_o)$ as the negative of the average weighted distance of $k$-nearest neighbors $\frac{1}{k}\sum_{i=1}^k -\parallel z_o-z_{(i)} \parallel*s_{(i)}$. The threshold $\lambda$ is chosen so that a high fraction of the ID validation data ($95\%$) can be correctly classified. The core component of \abbr is the memory augmentation in Algorithm~\ref{alg:main}, as explained in full detail below.

(1) \textbf{OOD sample selection and ID sample selection.} 
Using all online samples to augment the $\mathcal{M}$ is not desirable, as they add noise to the memory bank and may deteriorate performance. To avoid this issue, we use a selection margin $\gamma\geq 1$ to filter unreliable OOD samples. Only if the score of $x$ is smaller than $\gamma\lambda$, it will be a reliably detected OOD sample. Similarly, only if the score is higher than $\lambda/\gamma$, it will be a reliable ID sample.
(2) \textbf{Highlight the contribution of detected OOD samples.} Recall that our objective is to assign smaller scores to OOD samples than ID samples. However, given $x_o\in\mathcal{D}^{out}$, if its $k$-nearest neighbors are all OOD samples, they are highly similar and $-\parallel z_o-z_{i}\parallel$ will be large (small in magnitude), which is opposite to our goal. We thus introduce a large scale $\kappa\geq 1$ for reliably detected OOD samples to highlight their contributions\footnote{We use example to illustrate the importance of OOD scale $\kappa$. Assume  $\lambda=-1.0$ and $k=1$. Given $x_o=[2,2]\in\mathcal{D}^{out}$ and $\mathcal{M}=\{([1,1],1), ([0,0],1)\}$, we will get $S(x_o)=-1.4$. If $\mathcal{M}$ contains an OOD sample $([2.5,2.5],\kappa)$, we hope $x_o$ with OOD neighbors has a low score. In this case, the score will be $-0.707\kappa$. If $\kappa=1$, then $S(x_o)$ instead increases from $-1.4$ to $-0.7$ and will be identified as an ID sample due to the presence of OOD neighbors. We hence need a scale greater than $1$ to correct this phenomenon. $\kappa\rightarrow \infty$ means that once the $k$-nearest neighbors of $x_o$ contain OOD samples, then $x_o$ will be identified as OOD.
}.  (3) \textbf{Non-parametric adaptation}. After choosing reliably detected OOD samples and assigning them an appropriate scale, we augment the memory bank by including the selected feature vectors, i.e. $\mathcal{M}=\mathcal{M}\bigcup (z_o, \kappa)$. In contrast, existing TTA methods~\cite{sun2020test,wang2020tent,zhang2021test,zhang2022domain} require additional training or fine-tuning to add information from the detected OOD samples into models, leading to more computational overheads and making them less suitable for online setting.


\subsection{Discussion of the proposed setting and method}
\textbf{Difference to OOD detection with auxiliary OOD data.} This line of work regularizes the model during training. For example, the model is encouraged to give higher energies for out-of-distribution data~\cite{liu2020energy,ming2022poem,du2022unknown,katz2022training} or to give predictions with uniform distribution for out-of-distribution data points~\cite{lee2017training,hendrycks2018deep}. Recently, VOS~\cite{du2022towards} has tried to synthesize outliers and alleviate the need for auxiliary OOD data. However, the \lightgrey{real test/OOD distribution can be arbitrarily far from the auxiliary/synthetic OOD data}. In contrast with them, our work tries to directly utilize the online OOD instances during inference, not during training and outperforms all existing studies by a large margin.

\textbf{Can existing TTA methods be directly used for OOD detection tasks?}  (i) most existing online learning or TTA methods are primarily designed for image classification tasks. By contrast, our paper tackles the OOD detection problem, where \lightgrey{the label space of training data is different from the test OOD data}. This fundamental difference makes the OOD detection task inherently more challenging compared to domain generalization or traditional online learning. (ii) \lightgrey{existing TTA methods for image classification are not easily extendable to the OOD detection task} due to the discrepancy between training and testing targets. OOD detection requires distinguishing between known and unknown classes, and the lack of OOD samples during training makes it more complex than traditional image classification tasks. (iii) most TTA or online learning methods typically require adjustments to the model parameters, and we indeed design and compare various simple online learning baselines in our experiments. However, \lightgrey{OOD samples are usually scarce in real-world scenarios}, which can result in unstable gradients during online learning. Consequently, achieving stable and effective online learning in OOD detection becomes more challenging compared to traditional offline training. This is why our proposed AdaODD method, which leverages test-time samples to adapt the decision boundary, demonstrates advantages over methods that rely on gradient-based fine-tuning.

\section{Theoretical justification}\label{sec:app_theory}

\subsection{Log conditional
densities score function.}\label{sec:log}

To study how OOD samples influence the detection process, we partition the score function into two components, one dependent only on ID neighbors and the other dependent on selected test neighbors. Hereinafter, we treat the selected samples as low-confidence (LC) ID samples and the samples from naive ID distribution are high-confidence (HC) ID samples.  Because we do not have true labels of test samples, the LC ID sample set will contain both ID and OOD samples.\footnote{out-of-distribution samples are those which, when compared to the in-distribution, exhibit low confidence in their classification as part of the in-distribution.}. We show that, unlike previous methods that use HC ID neighbors and a fixed boundary, \abbr adjusts the boundary by the score from the LC sample information and prefers a lower false positive rate. Without access to OOD distribution, the typical decision rule is 
\vspace{-0.1cm}
\begin{align}
\mathbf{I}\{S(x|{D}^{HC})\geq \lambda\},\label{eq:xue:0}
\end{align}
where ${D}^{HC}$ is the original ID sample set. We treat samples with scores higher than $\lambda$ as ID. The potential drawback is that the value of $\lambda$ is typically relatively small so that the in-distribution detection maintains enough accuracy. If we can access the estimated LC ID information, we may use a combined decision rule as follows:
\begin{align}
\mathbf{I}\{S(x|{D}^{HC}, \hat{{D}}^{LC})\ge \lambda\},\label{eq:xue:1}
\end{align}
where $\hat{{D}}^{LC}$ is the low-confidence ID sample set during test. Intuitively, people usually hope the score function shows similar behavior as the probability function of ID, i.e., the higher score means a higher chance/probability that the test sample $x$ is from ID. Therefore, one of the ideal choices is to set the score function being proportional to log probability, i.e., $$S(x|{D}^{HC},\hat{{D}}^{LC}) \propto \log \mathbbm{P}(x\textrm{ from ID}|{D}^{HC},\hat{{D}}^{LC}).$$
We obtain the following equation through the maximum likelihood estimator analysis, 
\begin{align}
\small
S(x|{D}^{HC},\hat{{D}}^{LC})\propto \underbrace{\sum_{x_i\in{D}^{HC}}\log p^{hc}(x|x_i)}_{(a)} + \underbrace{\sum_{x_j\in\hat{{D}}^{LC}}\log p^{lc}(x|x_j)}_{(b)},\label{eq:xue:3}
\end{align}
where $p^{hc}(\cdot)$ and $p^{lc}(\cdot)$ are the conditional density/mass function of ID given a reference sample from HC sample set and LC sample set collected during test, respectively. 



\subsection{Nonparametric density estimator}\label{sec:app_density}

Until now, how to estimate the local density and LC ID sample set in \eqref{eq:xue:3} remains unclear. In this subsection, we apply KNN to approximate local densities~\cite{zhao2022analysis} and derive our decision rule. In section~\ref{sec:ood_sample}  we show how to estimate the LC ID sample set. To facilitate our discussion, we first consider an artificial setting in that we know the true in/out labels of the test samples, and every sample is normalized by its $l_2$ norm to project the representation onto the unit sphere. For a given test sample $x$, we first use cosine similarity to collect the $k$ closet samples from $x$ as ${D}_{KNN}(x)$ from ${D}^{in}\cup {D}^{test}$, where ${D}^{test}$ is test time sample set that follows some mixed distribution of ID/OOD. Without loss of generality, we assume ${D}_{KNN}(x)$ to contain $k_1$ and $k_2$ samples from HC/LC ID, respectively. We denote two disjoint datasets ${D}_{KNN}^{hc}(x)$ and ${D}_{KNN}^{lc}(x)$ containing the HC ID/LC ID samples in ${D}_{KNN}(x)$. Next, we make an assumption on the conditional densities $p^{hc}(x|x_i)$ and $p^{lc}(x|x_j)$ as follows:

{\it Assumption 1:} There exist two positive constants $\lambda_{hc} <\lambda_{lc}$ such that $p^{hc}(x|x_i)$ and $p^{lc}(x|x_j)$ are proportional to $\exp(-\lambda_{hc}\|x-x_i\|_2)$ and $\exp(-\lambda_{lc}\|x-x_j\|_2)$.\footnote{Here we assume the density function's tails performance being exponential. Via more sophisticated analysis, we may further relax the density being subgaussian/subexponential.}

 We use $\lambda_{hc}$ and $\lambda_{lc}$ to describe the influence of HC ID and LC ID samples. Let us consider a sample $x$ and a reference point $y$ that is close to $x$. 
 Intuitively, one may expect that $x$ has a higher chance of being an ID sample if $y$ is an HC ID sample and a lower chance if $y$ is believed from LC ID sample. In other words, for the same reference sample $y$, we should have $p^{hc}(x|y)> p^{lc}(x|y)$, which is achieved by setting $\lambda_{hc}<\lambda_{lc}$ in {\it Assumption 1}. During the testing stage, we can hardly ensure $y$ is from ID or OOD distribution. Therefore it would be safer to treat  $y$ as a reference sample that contributes less than a normal ID reference sample in assessing the chance of sample $x$ being ID.\footnote{If the exact OOD sample set is available in the test stage, the OOD detection task fall into the semi-supervised learning field, and we may use the log-likelihood-ratio~\cite{kawakita2013semi} as the score function, i.e., $S(x|{D}^{in},\hat{{D}}^{out}) \propto \log \left({\mathbbm{P}(x\textrm{ from ID}|{D}^{in})}/{\mathbbm{P}(x\textrm{ from OOD}|{D}^{out})}\right) 
 $. We detail the empirical results and comparison in section~\ref{sec:app_log}.} We then combine {\it Assumption 1} with ${D}^{hc}_{KNN}(x)$ and ${D}^{lc}_{KNN}(x)$ instead of ${D}^{hc}$ and $\hat{{D}}^{lc}$ in terms $(a)$ and $(b)$.  

\begin{equation}
\begin{aligned}
\tiny
&(a)\propto - \sum_{x_i\in{D}^{hc}_{KNN}(x)}\lambda_{hc}\|x-x_i\|_2\notag \\
&(b)\propto - \sum_{x_i\in{D}^{lc}_{KNN}(x)}\lambda_{lc}\|x-x_i\|_2\notag\\
&\Rightarrow (a) + (b) \propto -\sum_{x_i\in{D}_{KNN}(x)}\|x-x_i\|_2\cdot s_i\label{eq:xue:4}
\end{aligned}
\end{equation}

where we set $s_j = 1$ if $x_j$ is from HC ID and $s_j = \lambda_{lc}/\lambda_{hc}$ otherwise. Finally combine \eqref{eq:xue:4} with \eqref{eq:xue:3} and we end up with the decision rule used in {\bf Algorithm~\ref{alg1:decision} line 7}:
\begin{equation}
\begin{aligned}
\mathbf{I}\left\{-\sum_{x_j\in{D}_{KNN}} \|x-x_j\|_2 \cdot s_j \ge \lambda\right\}\label{eq:xue:5}
\end{aligned}
\end{equation}

Note that in the detection rule \eqref{eq:xue:5}, we require the knowledge on the distribution labels for all samples in ${D}_{KNN}$, which is inaccessible for samples from ${D}_{test}$ in practice. In this paper, we instead propose an estimation of distribution labels via \eqref{eq:xue:5} with a slightly larger $\lambda$. Since usually the ID sample size $N$ is much larger than the test sample size $n$, detecting extra ID samples only leads to marginal improvement. We may directly use $\mathcal{D}^{in}$ instead of all ID samples.

We use two scenarios in~\figurename~\ref{fig:motivation} to show the rationality of the decision rule \eqref{eq:xue:5}. 
\begin{figure}
    \centering
    \includegraphics[width=0.4\textwidth]{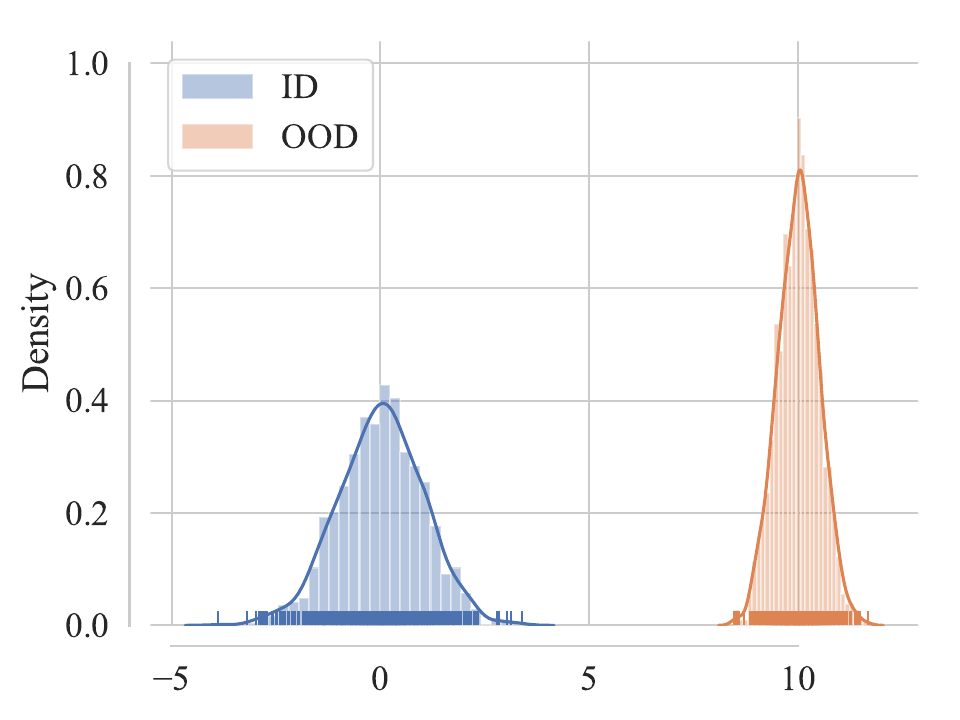}
    \caption{\small $\mathcal{D}^{out}=\mathcal{N}(10,1/2)$.}
        \label{fig:motivation1}
\end{figure}

\textit{Well separated ID/OOD in~\figurename~\ref{fig:motivation1}.} In this scenario, when $(a)$ in \eqref{eq:xue:4} is large enough (i.e., $x$ in a high-density regime of HC ID distribution), the density of LC ID distribution at $x$ must be very small, then $\log p^{lc}\approx 0$. In this case, the decision rule reduces to the classic detection rule \eqref{eq:xue:0}. On the other hand, when $x$ is from LC ID distribution, both $(a)$ and $(b)$ are expected to be small (large $\lambda_{lc}$ reduces the contribution of $p^{lc}$), and we can safely assign the OOD label while sacrificing minimal ID accuracy.

\textit{Overlapped ID/OOD in~\figurename~\ref{fig:motivation2}.} In this setting, we consider the ID/OOD overlap in some high-density regimes. The classic OOD detection rule \eqref{eq:xue:0} has a small threshold $\lambda$ and can lead to a large number of OOD samples being classified as ID samples. When applying the detection rule \eqref{eq:xue:4}, we measure the conditional density given both HC ID/LC ID sample sets. As long as one of the terms $(a)$ or $(b)$ is small enough, which corresponds to the situations where $x$ is far from the HC ID reference samples or has enough close estimated LC ID reference samples, we can still assign the sample $x$ with the OOD label. Our method, hence, yields a lower false positive rate.

\subsubsection{Estimation of OOD sample set.}\label{sec:ood_sample}
Next, we consider the estimation of the OOD sample set. Let denote ${D}_{KNN}^{lc,T}$ and ${D}_{KNN}^{lc,F}$ as the disjoint sample sets containing the true OOD samples and false OOD samples in ${D}_{KNN}^{lc}$. We will have the following separation:
\begin{equation}
\begin{aligned}
\tiny
&-\sum_{x_j\in{D}_{KNN}} s(x_j)\|x-x_j\|_2= \\
&-\sum_{x_j\in{D}_{KNN}^{lc,T}(x)}s(x_j)\|x-x_j\|_2  \\
&-\sum_{x_j\in{D}_{KNN}^{lc,F}(x)}s(x_j)\|x-x_j\|_2.  \\
\end{aligned}
\end{equation}
If we overestimate the OOD sample size by a margin, the corresponding $\sum_{x_j\in {D}_{KNN}^{lc}(x)}\|x-x_j\|_2s(x_j)$ will be overestimated and yield a lower AUC on ID/OOD detection. On the other hand, if we only collect the LC ID sample with enough confidence (e.g., the small enough value from the score function), the corresponding $\sum_{x_j\in {D}_{KNN}^{lc}(x)}\|x-x_j\|_2s(x_j)$ becomes smaller and the decision rule of \eqref{eq:xue:5} leans more on the high confidence ID samples. In the extreme case, the ${D}_{KNN}^{lc}(x)$ being empty set, the $\sum_{x_j\in {D}_{KNN}^{lc}(x)}\|x-x_j\|_2s(x_j)$ becomes 0 and the detection rule \eqref{eq:xue:5} reduces to the classic detection rule \eqref{eq:xue:0}. Therefore, only keeping the test sample with high confidence, as the detected OOD sample is safer. In this paper, we propose to use the following decision rule to collect the estimated OOD samples during the test stage ({\bf Algorithm~\ref{alg1:decision} line 9}):
\begin{align}
\mathbf{I}\left\{S(x_0|{D}^{hc}_{KNN}(x),\hat{{D}}^{lc}_{KNN}(x))<\lambda\cdot \gamma\right\},
\end{align}
where $\gamma\ge 1$ and $\hat{{D}}^{lc}_{KNN}(x)$ is the estimated OOD sample set and is empty at the beginning.

\section{Why shouldn't we use the log ratio test?}\label{sec:app_log}

As illustrated in Section~\ref{sec:app_density}, If the exact OOD sample set is available in the test stage, the OOD detection task falls into the semi-supervised learning field, and we should use the log-likelihood-ratio~\cite{kawakita2013semi} as the score function, i.e., $S(x|{D}^{in},\hat{{D}}^{out}) \propto \log \left({\mathbbm{P}(x\textrm{ from ID}|{D}^{in})}/{\mathbbm{P}(x\textrm{ from OOD}|{D}^{out})}\right)$. Consequently, the final score function will be 

\begin{equation}
\begin{aligned}
 S(x_o)=& \frac{1}{k}\sum_{i=1}^k \mathbf{1}\{y_i=OOD\}\parallel z_o-z_{(i)} \parallel s_{(i)}\\
 &-\frac{1}{k}\sum_{i=1}^k \mathbf{1}\{y_i=ID\}\parallel z_o-z_{(i)} \parallel s_{(i)}  
\end{aligned}
\label{equ:log_ratio}    
\end{equation}

As depicted in Table.~\ref{tab:log_ratio}, we observe notable OOD detection performance when a subset of OOD samples is accessible prior to the test phase. In our experiments, we preselect $10\%$ of the test set as OOD samples for log-ratio score calculation using \eqref{equ:log_ratio}. However, practical scenarios rarely offer the luxury of pre-access to OOD samples. In these cases, we adopt a strategy where each test sample is evaluated, and those with low confidence are stored as an estimated OOD distribution. This approach encounters challenges, primarily stemming from two factors:

1. \textbf{The Cold Start Problem}: At the outset of the test, the set of low-confidence samples is empty, leading to highly unstable and unreliable scores calculated in the early stages.

2. \textbf{Inaccurate Estimation}: As previously mentioned, it's impossible to guarantee that every sample in the low confidence set is genuinely an OOD sample. This introduces noise into the data, making \eqref{equ:log_ratio} susceptible to inaccuracies.

\begin{table}[th]
\caption{\textbf{OOD detection performance of log ratio test method} with ground truth OOD samples and only low-confidence samples during the test.}\label{tab:log_ratio}

\centering
\begin{tabular}{ccccc}
\toprule
\multirow{2}{*}{Dataset} & \multicolumn{2}{c}{With GT OOD samples} & \multicolumn{2}{c}{With LC ID samples} \\ \cmidrule{2-3}\cmidrule{4-5}
 & FPR95 & AUROC & FPR95 & AUROC \\ \hline
SVHN & 0.1 & 99.96 & 99.85 & 5.01 \\
LSUN & 0.57 & 99.86 & 99.81 & 1.57 \\
iSUN & 0.17 & 99.92 & 99.76 & 0.27 \\
dtd & 2.82 & 99.35 & 99.67 & 0.44 \\ \hline
AVG & 0.91 & 99.77 & 99.77 & 1.82 \\ \bottomrule
\end{tabular}
\end{table}
In contrast, our proposed method, referred to as \abbr, effectively addresses these issues. In conclusion, theoretical results in Section~\ref{sec:app_theory} provide insights into the underlying mechanisms of our approach, and by understanding these principles, researchers can better utilize and adopt our framework for various applications. Besides, we show that, unlike previous methods that use ID neighbors and a fixed boundary, \abbr adjusts the boundary by the score from the estimated OOD information and prefers a lower false positive rate whether the ID and OOD distributions are well-separated and overlapped.


\section{Experiments}\label{sec:exp}

\begin{table*}[]
\caption{\textbf{Results on small benchmarks}, where the ID dataset is CIFAR-10. All methods are based on a model trained on ID data. $\uparrow$ indicates larger values are better and vice versa.} \label{tab:cifar}
\resizebox{\textwidth}{!}{%
\begin{tabular}{lcccccccccccc}
\toprule
\multicolumn{1}{c}{\multirow{2}{*}{Method}} & \multicolumn{2}{c}{\textbf{SVHN}} & \multicolumn{2}{c}{\textbf{LSUN}} & \multicolumn{2}{c}{\textbf{iSUN}} & \multicolumn{2}{c}{\textbf{Texture}} & \multicolumn{2}{c}{\textbf{Place365}} & \multicolumn{2}{c}{\textbf{Average}} \\
\multicolumn{1}{c}{} & \textbf{FPR $\downarrow$} & \textbf{AUROC $\uparrow$} & \textbf{FPR $\downarrow$} & \textbf{AUROC $\uparrow$} & \textbf{FPR $\downarrow$} & \textbf{AUROC $\uparrow$} & \textbf{FPR $\downarrow$} & \textbf{AUROC $\uparrow$} & \textbf{FPR $\downarrow$} & \textbf{AUROC $\uparrow$} & \textbf{FPR $\downarrow$} & \textbf{AUROC $\uparrow$} \\
\midrule
\multicolumn{13}{c}{\textbf{Without Contrastive Learning}} \\
MSP & 59.66 & 91.25 & 45.21 & 93.80 & 54.57 & 92.12 & 66.45 & 88.50 & 62.46 & 88.64 & 57.67 & 90.86 \\
ODIN & 20.93 & 95.55 & 7.26 & 98.53 & 33.17 & 94.65 & 56.40 & 86.21 & 63.04 & 86.57 & 36.16 & 92.30 \\
Energy & 54.41 & 91.22 & 10.19 & 98.05 & 27.52 & 95.59 & 55.23 & 89.37 & 42.77 & 91.02 & 38.02 & 93.05 \\
GODIN & 15.51 & 96.60 & \textbf{4.90} & 99.07 & 34.03 & 94.94 & 46.91 & 89.69 & 62.63 & 87.31 & 32.80 & 93.52 \\
Mahalanobis & 9.24 & 97.80 & 67.73 & 73.61 & 6.02 & 98.63 & 23.21 & 92.91 & 83.50 & 69.56 & 37.94 & 86.50 \\
KNN & 24.53 & 95.96 & 25.29 & 95.69 & 25.55 & 95.26 & 27.57 & 94.71 & 50.90 & 89.14 & 30.77 & 94.15 \\ \Gray
\abbr & \textbf{2.17} & \textbf{99.35} & {5.80} & \textbf{98.90} & \textbf{4.90} & \textbf{98.89} & \textbf{15.82} & \textbf{96.74} & \textbf{7.50} & \textbf{97.31} & \textbf{7.24} & \textbf{98.24} \\ \midrule
\multicolumn{13}{c}{\textbf{With Contrastive Learning}} \\
CSI & 37.38 & 94.69 & 5.88 & 98.86 & 10.36 & 98.01 & 28.85 & 94.87 & 38.31 & 93.04 & 24.16 & 95.86 \\
SSD+ & \textbf{1.51} & 99.68 & 6.09 & 98.48 & 33.60 & 95.16 & 12.98 & 97.70 & 28.41 & 94.72 & 16.52 & 97.15 \\
KNN & 2.42 & 99.52 & 1.78 & 99.48 & 20.06 & 96.74 & 8.09 & 98.56 & 23.02 & 95.36 & 11.07 & 97.93 \\ \Gray
\abbr & {1.79} & \textbf{99.71} & \textbf{1.79} & \textbf{99.67} & \textbf{6.81} & \textbf{98.97} & \textbf{6.97} & \textbf{98.92} & \textbf{10.19} & \textbf{97.53} & \textbf{5.51} & \textbf{98.96} \\ \bottomrule
\end{tabular}%
}
\end{table*}

\begin{table*}[]
\caption{\textbf{Results on hard OOD detection tasks}, where the ID dataset is CIFAR-10. All methods are based on a model trained by supervised contrastive loss. $\uparrow$ indicates larger values are better.} \label{tab:cifar_hard}
\centering
\resizebox{0.98\textwidth}{!}{%
\begin{tabular}{ccccccccccc}
\toprule
\multirow{2}{*}{Method} & \multicolumn{2}{c}{\textbf{LSUN-FIX}} & \multicolumn{2}{c}{\textbf{ImageNet-FIX}} & \multicolumn{2}{c}{\textbf{ImageNet-Resize}} & \multicolumn{2}{c}{\textbf{CIFAR-100}} & \multicolumn{2}{c}{\textbf{Avg}} \\
 &  \textbf{FPR $\downarrow$} & \textbf{AUROC $\uparrow$} & \textbf{FPR $\downarrow$} & \textbf{AUROC $\uparrow$} & \textbf{FPR $\downarrow$} & \textbf{AUROC $\uparrow$} & \textbf{FPR $\downarrow$} & \textbf{AUROC $\uparrow$} & \textbf{FPR $\downarrow$} & \textbf{AUROC $\uparrow$} \\ \midrule
SSD+ & 29.86 & 94.62 & 23.26 & 93.19 & 45.62 & 89.72 & 45.50 & 87.34 & 36.06 & 91.22 \\
KNN & 21.52 & 96.51 & 25.92 & 95.71 & 30.16 & 95.08 & 38.83 & 92.75 & 29.11 & 95.01 \\
\abbr & \textbf{12.04} & \textbf{98.16} & \textbf{22.81} & \textbf{96.45} & \textbf{12.48} & \textbf{97.75} & \textbf{34.73} & \textbf{93.46} & \textbf{20.52} & \textbf{96.46} \\ \bottomrule
\end{tabular}%
}
\end{table*}

In this section, we show that (1) \abbr attains superior performances on small and large-scale benchmarks; (2) we design several strong TTA baselines, which are all shown be inferior to \abbr; (3) \abbr performs well under different OOD detection settings, and (4) how to select the best hyper-parameters for different benchmarks. Besides, we evaluate \abbr on various real-world OOD detection settings and further explore the effect of covariate shift on the detection performance.

\textbf{Benchmarks.} We study OOD detection with the following two settings: (1) \textbf{Small Benchmarks that are routinely used.} We use $50,000$ training images and $10,000$ test images of the CIFAR benchmarks for model training. The model is evaluated on common OOD datasets: SVHN~\cite{netzer2011reading}, LSUN~\cite{yu2015lsun}, Places365~\cite{zhou2017places}, Textures~\cite{cimpoi2014describing}, and iSUN~\cite{xu2015turkergaze} and more chanllenging OOD datasets: LSUN-FIX~\cite{yu2015lsun}, ImageNet-FIX~\cite{deng2009imagenet}, ImageNet-Resize~\cite{deng2009imagenet}, and CIFAR-100.  (2) \textbf{Large-scale ImageNet Task~\cite{huang2021mos}} that contains a widespread range of domains such as fine-grained images, scene images, and textural images. The ImageNet~\cite{deng2009imagenet} training model is evaluated on four OOD datasets: Places365~\cite{zhou2017places}, Textures~\cite{cimpoi2014describing}, iNaturalist~\cite{van2018inaturalist}, and SUN~\cite{xiao2010sun}. (3) \textbf{Other evaluation settings}. We evaluate \abbr under different test distributions, such as (a) OOD samples are from different OOD distributions sequentially, namely, the algorithm is evaluated and adapted on a series of OOD datasets; (b) the test OOD distribution is a mixture of multiple OOD distributions rather than coming from the same distribution; (c) the test distribution is a mixture of ID and OOD distributions.

\textbf{Baselines.} We incorporate several baselines that derive OOD scores from a model trained with common softmax cross-entropy loss, such as MSP~\cite{hendrycks2016baseline}, ODIN~\cite{liang2018enhancing}, Mahalanobis~\cite{lee2018simple}, Energy~\cite{liu2020energy}, GODIN~\cite{hsu2020generalized}, KNN~\cite{sun2022out}. In addition, there are several baselines that benefit from contrastive learning, such as CSI~\cite{tack2020csi}, SSD+~\cite{sehwag2021ssd}, and KNN~\cite{sun2022out}.

\textbf{Evaluation metrics} include (1) the false positive rate (FPR95) of OOD samples when the true positive rate of ID is at $95\%$. (2) the area under the receiver operating characteristic (AUROC), which tells us whether our model is able to correctly rank examples. (3) the inference time per image that is averaged from all test images in milliseconds.

\subsection{Experimental settings.}\label{sec:data_detail}
\begin{algorithm}[h]
   \caption{AdaODD for the synthetic dataset in \figurename~\ref{fig:motivation2}}
   \label{alg:main2}
{\bfseries Input:} ID dataset ${D}^{in}$, accessible OOD dataset ${D}^{out}=\{x_i\}_{i=1}^m$, test sample $x_{o}$, threshold $\lambda$, $k=100$ for KNN selection, sample scale $\kappa=100.$.\\
{\bfseries Initialize} the memory bank with scales $\mathcal{M}=\{x_i, s_i:=1\}_{x_i\in {D}^{in}}\bigcup \{x_i, s_i:=\kappa\}_{x_i\in {D}^{out}}$.\\
{\bfseries Inference stage for $x_{o}$.}\\
 Nearest neighbors $B=\left((x_{(1)},s_{(1)}),...,(x_{k)},s_{(k)})\right)$. \\
 Calculate score $S(x_o)=\frac{1}{k}\sum_{i=1}^k -\parallel x_o-x_{(i)} \parallel*s_{(i)}$.\\
Make desion by $\mathbf{1}\{S(x_o)\geq \lambda\}$\\
\end{algorithm}

\textbf{Experimental setting for the synthetic dataset in \figurename~\ref{fig:motivation2}.} We sample $5000$ ID instances from $\mathcal{D}^{in}$, which serve as the KNN memory bank in Algorithm~\ref{alg:main2}. $5000+m$ test OOD instances are sampled from $\mathcal{D}^{out}$, where $5000$ instances are used for inference and other $m$ OOD instances are accessible before test. Specifically, we add these accessible instances to the memory and assign them large weights $\kappa$. Algorithm~\ref{alg:main2} is a simplified version of the proposed \abbr in Algorithm~\ref{alg:main}. The main difference is that Algorithm~\ref{alg:main2} has accessible OOD samples before the inference stage but Algorithm~\ref{alg:main} has only unlabeled test data. Using Algorithm~\ref{alg:main2} enables us to focus on studying how the provided OOD samples influence the detection result. We repeat the experiment $20$ times, showing the mean and the error bar of the false positive rate.

\textbf{Hyper-parameters details} We use $k= 50$ for CIFAR-10, $k=200$ for CIFAR-100 and $k=1000$ for ImageNet, which is selected from $k=\{1,10,20,50,100,200,500,1000,3000,5000\}$ using the validation method in~\cite{hendrycks2018deep}. The margin $\gamma$ is $1.0, 1.5$ for ImageNet and CIFAR-10, respectively, where $\gamma$ is chosen from $\{1.0,1.1,....,2.0\}$. The scale of out-of-distribution samples $\kappa$ is set to $10.0,5.0$ for ImageNet and CIFAR-10, respectively, which is selected from $\{1.0,2.0,...,10.0,100.0\}$, and we will detail the effect of all hyperparameters in the ablation studies.

\textbf{Implementation details.} All of our experiments are based on Pytorch~\cite{paszke2019pytorch}. The $k$-nearest neighbor search is based on faiss~\cite{johnson2019billion} library, where the distance is calculated by Euclidean distance. We conduct all the experiments on a machine with Intel R Xeon (R) Platinum 8163 CPU @ 2.50GHZ, 32G RAM, and Tesla-V100 (32G)x4 instances.

\textbf{License.} All the assets (i.e., datasets and the codes for baselines) we use the MIT license containing a copyright notice and this permission notice shall be included in all copies or substantial portions of the software.

\subsection{Evaluation and Ablations on Small Out-of-distribution Detection Benchmarks}


\textbf{Experimental details.} For CIFAR-10, we use ResNet-18~\cite{he2016deep} as the backbone. We conduct two different settings considering the ID training process: (1) The training loss is cross-entropy loss. (2) The training loss is supervised contrastive loss~\cite{khosla2020supervised} with temperature $\tau=0.1$.

\begin{table}[t]
\centering
\caption{\textbf{FPR95 of different TTA methods on small benchmarks}. The speed is the inference time in ms.}\label{tab:tta}

\resizebox{0.99\linewidth}{!}{%
\begin{tabular}{ccccccc}
\toprule
 & SVHN & LSUN & iSUN & Texture & Avg & Speed \\ \midrule
Energy & 54.41 & 10.19 & 27.52 & 55.23 & 36.84 & 4.81 \\
TTA Energy & 28.95 & 50.00 & 16.67 & 60.00 & 38.91 & 24.50 \\
Mahalanobis & 59.20 & 11.01 & 54.47 & 31.61 & 39.07 & 21.82 \\
TTA Mahalanobis & 50.00 & 18.42 & 39.28 & 27.64 & 33.84 & 50.98 \\
KNN & 24.53 & 25.29 & 25.55 & 27.57 & 25.74 & 4.97 \\
TTA KNN & 26.00 & 14.38 & 17.95 & 24.70 & 20.76 & 24.85 \\  \Gray
AdaODD & \textbf{2.17} & \textbf{5.80} & \textbf{4.90} & \textbf{15.82} & \textbf{7.17} & 5.84
\\ \bottomrule
\end{tabular}%
}
\end{table}
\textbf{\abbr outperforms all baselines by a large margin.} Results are shown in Table~\ref{tab:cifar}, where the proposed \abbr reduces the average FPR95 of previous best model from $\mathbf{30.77}$ to $\mathbf{7.24}$. Note that \abbr achieves superior results in large OOD datasets, such as the Places365~\cite{zhou2017places} dataset, where the best FPR95 of the baseline models is $\mathbf{42.77}$ but \abbr achieves $\mathbf{7.5}$. Taking into account the models trained by the loss of supervised contrast learning (SupCon), all methods, including CSI~\cite{tack2020csi}, SSD+~\cite{sehwag2021ssd}, and KNN~\cite{sun2022out} achieve superior results compared to models trained by cross-entropy loss. The proposed nonparametric test-time adaptation method is agnostic to the training procedure and is compatible with models trained under different losses. Table~\ref{tab:cifar} shows that the FPR95 of \abbr with SupCon loss ($\mathbf{5.51}$) is only half of the previous best result ($\mathbf{11.07}$). 

\textbf{\abbr outperforms strong TTA baselines.} We modify several baselines into TTA by proposing some specific training strategies during test. In particular, given a detected OOD sample $x_{out}$ and a set of ID samples $x_{in}\sim\mathcal{D}_{in}$, we add the following training loss during inference as follows (1) \textbf{TTA Energy~\cite{liu2020energy}}: $L_{energy}=\max(0,E(x_{in}-m_{in}))^2+\max(0,m_{out}-E(x_{out}))^2$, where $m_{out}$ is the separate margin hyperparameter for the two squared hinge loss and $E$ is the energy function. Namely, we penalize the out-of-distribution samples with energy lower than the margin parameter $m_{out}$. The  $m_{in}$ is chosen from $[-3, -5, -7]$ and $m_{out}$ is chosen from $[-15,-19,-23,-27]$ as~\cite{liu2020energy}. (2) \textbf{TTA Mahalanobis~\cite{lee2018simple}}, and \textbf{TTA KNN~\cite{sun2022out}}. We use metric learning to enlarge the margin between ID and OOD representations. Specifically, for the anchor sample $x_{out}$,we sample $x_{in}\sim\mathcal{D}_{in}$ as the negative sample, a random sampled $\bar{x}_{out}$ as the positive sample. Finally, we optimize the triplet loss $L_{tri}=\max(d(x_{out},\bar{x}_{out})-d(x_{out},x_{in})+m,0)$, where $m$ is the margin and $d$ is the distance (L-2 norm by default). The results in Table~\ref{tab:tta} demonstrate that all of the strong baselines designed are better than their offline counterparts, which confirms the significance of test time adaptation. However, they are still outperformed by \abbr. It is worth noting that all of these methods require retraining the model when an OOD sample is detected, resulting in computational burdens and limiting their practicality. Additionally, there are some cases where the baseline method without any TTA has a lower FPR95: The reason is that test-time training is not always beneficial during inference, since the model is trained using only one data instance instead of a batch of data, which leads to highly noisy gradients that could potentially harm the model optimization process. In contrast. AdaODD doesn’t need a gradient update and is more stable.

\textbf{\abbr performs well on more challenging OOD tasks.} We follow \cite{sun2022out,tack2020csi} and evaluate \abbr on several hard OOD datasets: LSUN-FIX~\cite{yu2015lsun}, ImageNet-FIX~\cite{deng2009imagenet}, ImageNet-Resize~\cite{deng2009imagenet}, and CIFAR-100, where the OOD samples are particularly challenging to detect. Table~\ref{tab:cifar_hard} shows that even with models trained with SupCon loss, the baselines SSD+~\cite{sehwag2021ssd} and KNN~\cite{sun2022out} cannot achieve a low false positive rate as in Table~\ref{tab:cifar}, revealing the difficulty of the tasks. However, \abbr reduces the FPR95 by $\mathbf{15.58}$ and $\mathbf{8.59}$ compared to SSD+~\cite{sehwag2021ssd} and KNN~\cite{sun2022out} respectively.

\begin{figure}[h]
    \centering
\minipage{0.49\linewidth}%
  \includegraphics[width=\linewidth]{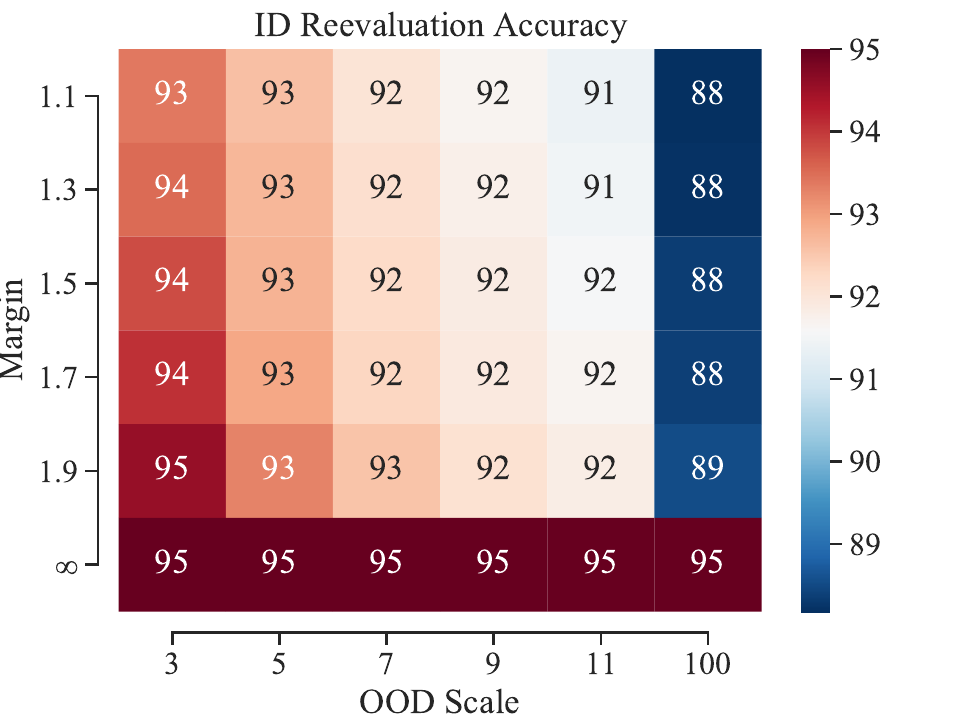}
\endminipage\hfill
\minipage{0.49\linewidth}
  \includegraphics[width=\linewidth]{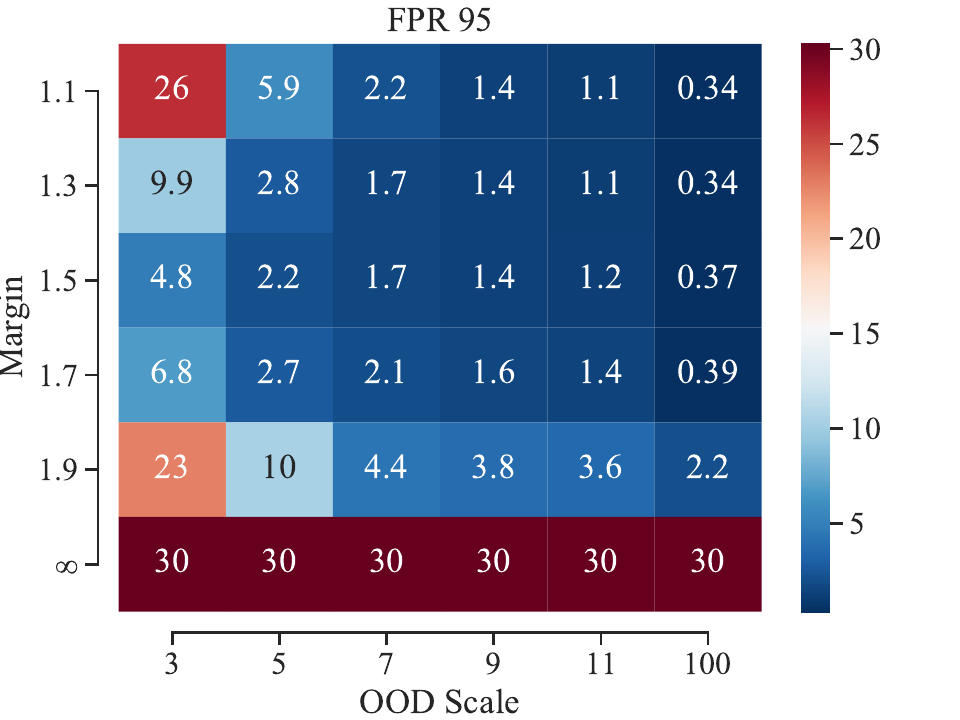}
\endminipage\hfill
\caption{\textbf{Ablation studies of OOD selection margin $\gamma$ and sample scale $\kappa$}. (a) Reevaluation accuracy of ID samples and (b) FPR95 of OOD samples in the SVHN~\cite{netzer2011reading} dataset.}\label{fig:cifar_magin}\label{fig:cifar_magina}\label{fig:cifar_maginb}
\end{figure}

\begin{table*}[t]
\caption{\textbf{Results on large-scale benchmarks}, where the ID dataset is ImageNet. All methods are based on a model trained on ID data only. $\uparrow$ indicates larger values are better.} \label{tab:imagenet}

\resizebox{\textwidth}{!}{%
\begin{tabular}{@{}lccccccccccc@{}}
\toprule
\multirow{2}{*}{\textbf{Method}} & \multirow{2}{*}{\textbf{\begin{tabular}[c]{@{}c@{}}Inference \\ time (ms)\end{tabular}}} & \multicolumn{2}{c}{\textbf{iNaturalist}} & \multicolumn{2}{c}{\textbf{SUN}} & \multicolumn{2}{c}{\textbf{Places}} & \multicolumn{2}{c}{\textbf{Textures}} & \multicolumn{2}{c}{\textbf{Average}} \\ 
 & & \textbf{FPR $\downarrow$} & \textbf{AUROC $\uparrow$} & \textbf{FPR $\downarrow$} & \textbf{AUROC $\uparrow$} & \textbf{FPR $\downarrow$} & \textbf{AUROC $\uparrow$} & \textbf{FPR $\downarrow$} & \textbf{AUROC $\uparrow$} & \textbf{FPR $\downarrow$} & \textbf{AUROC $\uparrow$} \\ \midrule
MSP & 7.04 & 54.99 & 87.74 & 70.83 & 80.86 & 73.99 & 79.76 & 68.00 & 79.61 & 66.95 & 81.99 \\
ODIN & 7.05 & 47.66 & 89.66 & 60.15 & 84.59 & 67.89 & 81.78 & 50.23 & 85.62 & 56.48 & 85.41 \\
Energy & 7.04 & 55.72 & 89.95 & 59.26 & 85.89 & 64.92 & 82.86 & 53.72 & 85.99 & 58.41 & 86.17 \\
GODIN & 7.04 & 61.91 & 85.40 & 60.83 & 85.60 & 63.70 & 83.81 & 77.85 & 73.27 & 66.07 & 82.02 \\
Mahalanobis & 35.83 & 97.00 & 52.65 & 98.50 & 42.41 & 98.40 & 41.79 & 55.80 & 85.01 & 87.43 & 55.47 \\
KNN & 10.31 & 59.00 & 86.47 & 68.82 & 80.72 & 76.28 & 75.76 & 11.77 & 97.07 & 53.97 & 85.01 \\ \Gray
\abbr & 10.31 & \textbf{4.48} & \textbf{98.85} & \textbf{18.70} & \textbf{93.76} & \textbf{33.59} & \textbf{87.81} & \textbf{7.13} & \textbf{98.22} & \textbf{15.97} & \textbf{94.66} \\ \bottomrule
\end{tabular}%
}

\end{table*}
\begin{figure}[h]
    \centering
    \minipage{0.49\linewidth}%
    \includegraphics[width=\linewidth]{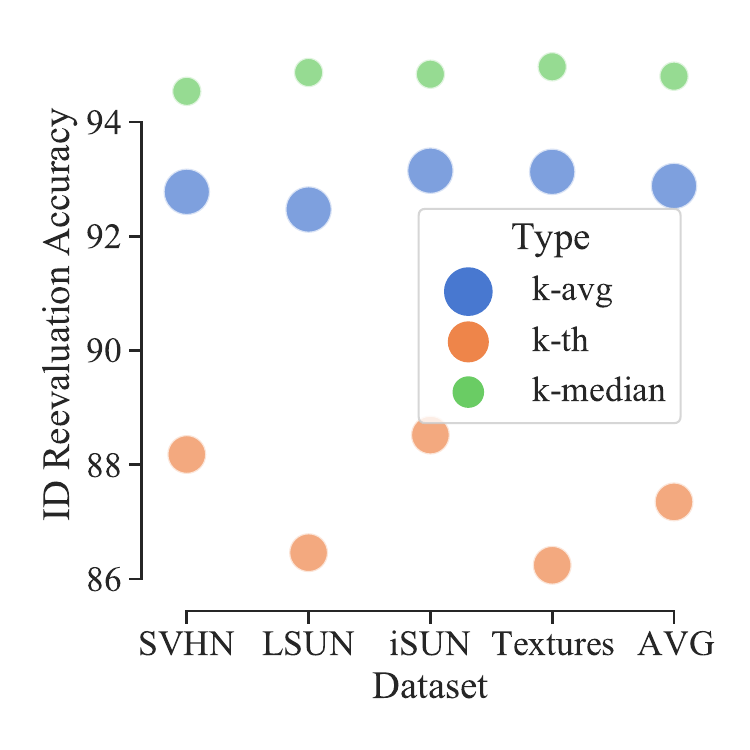}
\endminipage\hfill
\minipage{0.49\linewidth}
  \includegraphics[width=\linewidth]{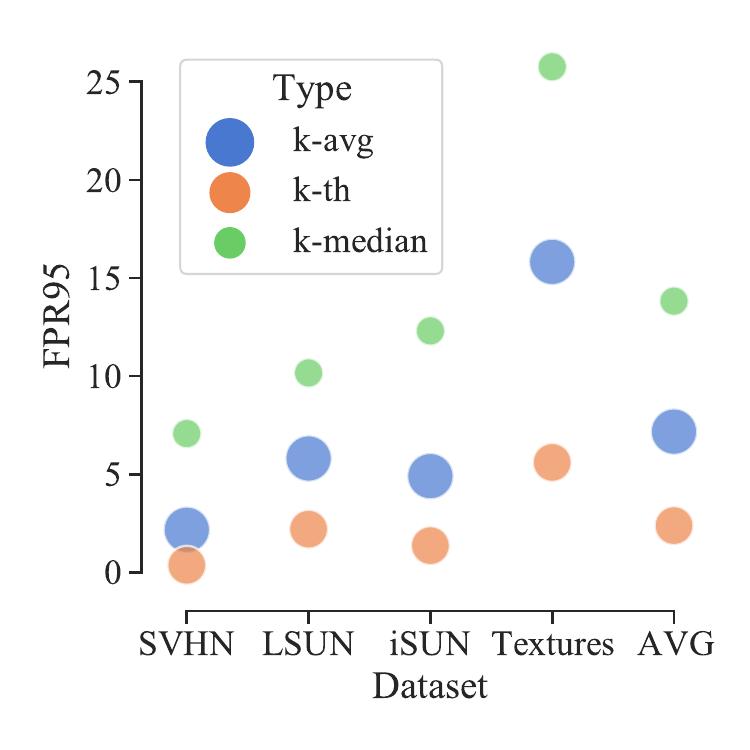}
\endminipage
\caption{\textbf{Ablation of using $k$-avg, $k$-th, $k$-median nearest neighbor distance} considering different OOD datasets on small OOD detection benchmark. (a) Reevaluation, (b) FPR.}\label{fig:cifar_avg}\label{fig:cifar_avg_in}\label{fig:cifar_avg_out}
\end{figure}
\textbf{The effect of test-time adaptation for ID data.} One main concern is that as adaptation progresses, the ID samples will be harder to identify because more outliers are added to the memory bank and large scales are assigned. To address this concern, we reevaluate the adapted model in the ID samples, which can test whether the adapted model can still identify the ID examples successfully. For example, in~\figurename~\ref{fig:cifar_magina}, we first evaluate and adapt the algorithm to SVHN dataset. After the adaptation, we evaluate the adapted model on ID samples, namely using the augmented memory bank and calculate the score by $\frac{1}{k}\sum_{i=1}^k -\parallel z_o-z_{(i)} \parallel*s_{(i)}$ in Algorithm~\ref{alg:main}. The evaluation metric is termed \textbf{ID reevaluation accuracy}. To conclude, the results show that the adaptation process only slightly reduces the accuracy of the ID samples but achieves a significant benefit for the detection of the OOD samples compared to the non-adapted model, whose FPR95 is $30.28$. 

\textbf{How to choose the best $\gamma$ and $\kappa$?}  We study the effect of selection margin $\gamma$ and sample scale $\kappa$ in \figurename~\ref{fig:cifar_magin}. We observe that: (i) A selection margin $\gamma$ too small or too large leads to inferior OOD detection results (\figurename~\ref{fig:cifar_maginb}). The reason is that the small $\gamma$ brings many noisy samples that may not be OOD instances. Large $\gamma$ cannot utilize the benefit of test-time adaptation because only a small part OOD samples satisfy the condition, namely $score>\lambda*\gamma$. Hence, larger $\gamma$ attains inferior OOD detection results but retains better ID detection accuracy. (ii) A larger scale $\kappa$ always leads to a better OOD detection performance because all OOD samples are of large weight. In the extreme case, namely $\kappa=\infty$, given a test sample $x^*$, once there is an OOD sample in its $k$ nearest neighbors, $x^*$ will attain a very low score and be decided as an OOD sample. At the same time, using a large $\kappa$ leads to a worse ID detection performance (\figurename~\ref{fig:cifar_magina}). We choose $\kappa=1.5$ and $\gamma=5.0$ by default so that the adapted model retains a comparable ID detection accuracy to models without TTA.

\begin{figure}[h]
    \centering
\minipage{0.49\linewidth}
  \includegraphics[width=\textwidth]{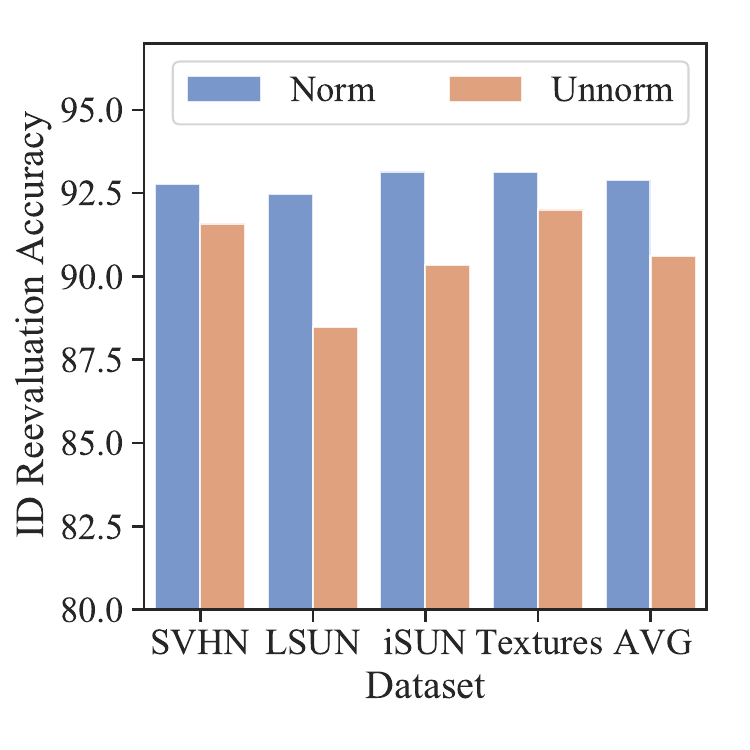}
\endminipage\hfill
\minipage{0.49\linewidth}
\includegraphics[width=\textwidth]{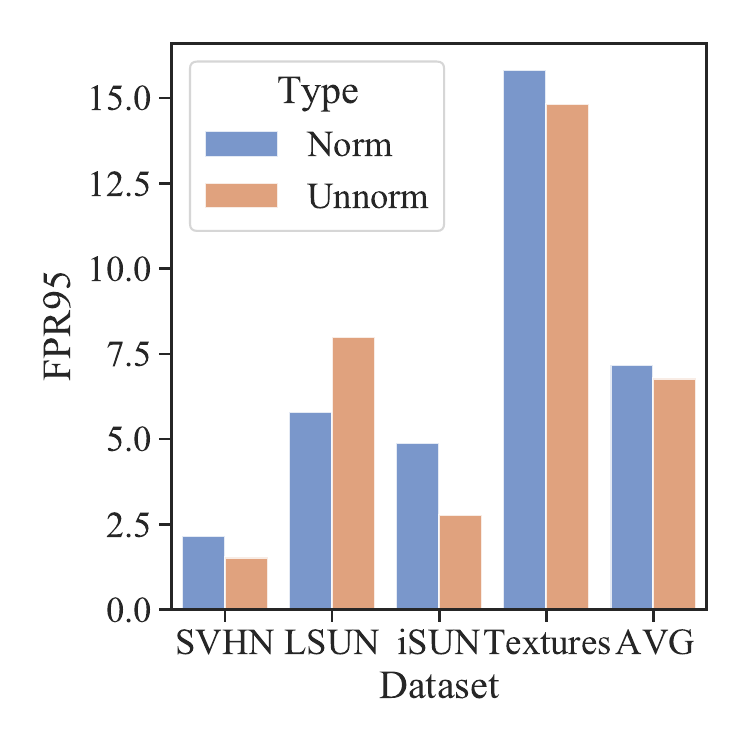}
\endminipage
\caption{\textbf{ Ablation of the effect of using normalization in the features vs. without normalization} (a) Reevaluation accuracy of ID samples and (b) FPR95 of OOD samples in the LSUN~\cite{yu2015lsun}.
}\label{fig:cifar_norm}\label{fig:cifar_norm_in}\label{fig:cifar_norm_out}
\end{figure}

\textbf{Do we need to normalize features before $k$-nearest neighbors search?} Note that in Algorithm~\ref{alg:main}, before searching the nearest neighbors, we need to normalize the feature vectors by $h_\theta(x) /\parallel h_\theta(x) \parallel_2$. \figurename~\ref{fig:cifar_norm} shows that the normalization process is not necessary for small-scale datasets, and the false positive rate of \abbr without normalization is slightly better than that with normalization. However, without feature normalization, the ID reevaluation accuracy will be greatly reduced. Therefore, in this section, we use feature normalization for experiments as default. The normalization process is more important for large scale benchmarks, which will be discussed in the next subsection.

\textbf{How to combine the score from $k$ nearest neighbors?} Given a test sample $x_o$ and its $k$-nearest neighbors, we can calculate $k$ score by decreasing order $\left(-\parallel z_o-z_{(1)} \parallel*s_{(1)},...,-\parallel z_o-z_{(k)} \parallel*s_{(k)}\right)$. Here, we compare three different scoring calculation methods: the $k$-th nearest neighbor score ($-\parallel z_o-z_{(k)} \parallel*s_{(k)}$), the average of $k$ nearest neighbor scores ($\frac{1}{k}\sum_{i=1}^k-\parallel z_o-z_{(i)} \parallel*s_{(i)}$), and the median of $k$ nearest neighbor scores ($-\parallel z_o-z_{\lfloor k/2\rfloor} \parallel*s_{\lfloor k/2\rfloor}$), where $\lfloor k/2\rfloor$ is the value of rounding down $k/2$. \figurename~\ref{fig:cifar_avg} shows that $k$ -th score achieves the lowest false positive rate; however, unlike $k$ -avg and $k-$ median scores, the model with $k$-th score cannot identify ID samples successfully, where the precision is usually below $90\%$. $k-$median score achieves high ID accuracy but poor OOD detection results compared to its variants. We choose $k-$avg score as default because it makes a good trade-off between ID reevaluation accuracy and OOD detection performance.

\subsection{Evaluation and Ablations on Large-scale Out-of-distribution Detection Benchmarks}

\begin{table}[]
\caption{\textbf{FPR95 on large-scale benchmarks} with the ViT-B/16 backbone. The ID dataset is ImageNet.} \label{tab:vit}

    \centering
    \resizebox{\linewidth}{!}{%
\begin{tabular}{@{}lccccc@{}}
\toprule
 & \textbf{iNaturalist} & \textbf{SUN} & \textbf{Places} & \textbf{Textures} & \textbf{Avg} \\ \midrule
Mahalanobis & 17.56 & 80.51 & 84.21 & 70.51 & 63.19 \\
KNN & 7.30 & 48.41 & 56.46 & 39.91 & 38.01 \\
\abbr & \textbf{2.37} & \textbf{13.39} & \textbf{24.69} & \textbf{24.01} & \textbf{16.11} \\ \bottomrule
\end{tabular}}
\end{table}

\textbf{Experimental details.} We use a ResNet-50~\cite{he2016deep} backbone by default and train the model by cross-entropy loss. The ImageNet training set~\cite{deng2009imagenet} has $12,800,000$ images with resolution $224\times 224$, making detection tasks more challenging than small-scale experiments. All models are loaded directly from Pytorch~\cite{paszke2019pytorch} and the feature dimension is $2,048$.

\textbf{\abbr reduces the previous best FPR from $53.97$ to $15.97$.} As shown in Table~\ref{tab:imagenet}, utilizing online test samples improves the detection performance by a large margin. Especially on the iNaturalist dataset, FPR is reduced to $1/10$ of the SOTA model. In addition to the surprising performance of OOD detection, \abbr retains a comparable inference time compared to baselines. Besides, \textbf{\abbr benefits from large backbones}. Table~\ref{tab:vit} shows that, with more powerful representations of the ViT-B/16 backbone, the OOD detection performance of \abbr will be further improved. Although some of the baselines can also benefit from better backbones, their performance still has a large gap compared to \abbr.

\textbf{Computational complexity.} 
We have leveraged FAISS to mitigate computational overhead and enhance efficiency. A comprehensive analysis of inference times is presented in Table~\ref{tab:imagenet}, underscoring the efficiency of our KNN-based method in comparison to various existing approaches, including TTA methods. Previous research~\cite{hardt2023test} has demonstrated that even with considerably larger datasets, the retrieval process remains swift. An additional concern pertains to whether the indexing data structure of FAISS necessitates updates each time a test sample is added to the memory bank, and whether this process incurs a substantial computational burden. Notably, the cost of updating the indexing data structure is substantially lower than that of the search process. Moreover, in practical, there's no requirement to update the indexing data structure for every individual test instance.

\begin{figure}[h]
    \centering
    \minipage{0.45\textwidth}%
  \includegraphics[width=\linewidth]{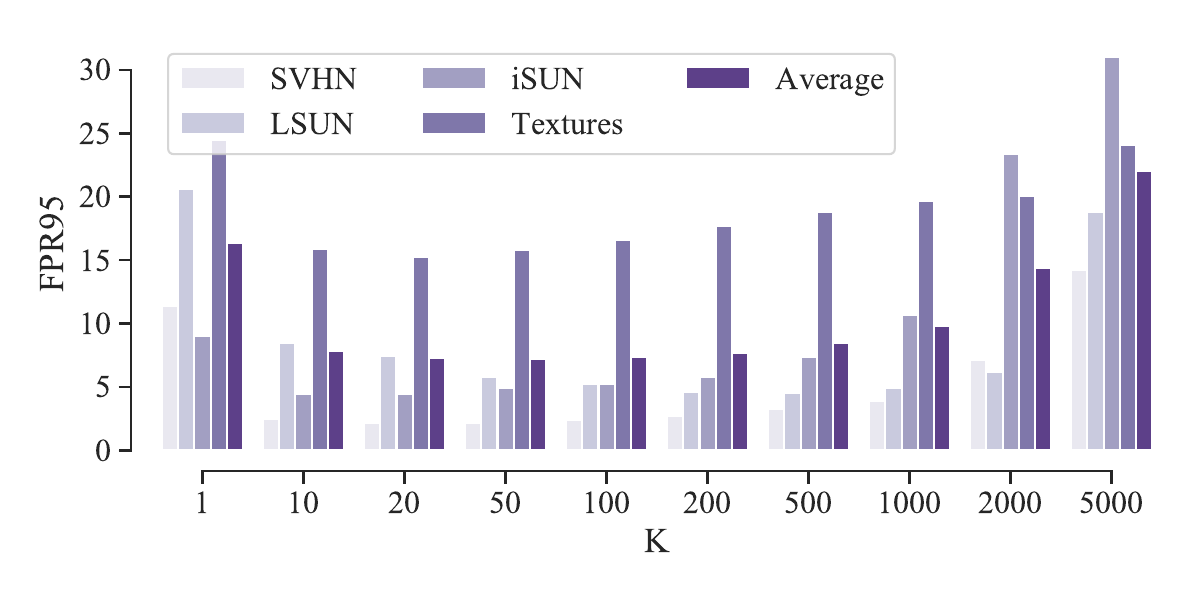}
\endminipage\hfill
\minipage{0.45\textwidth}
 \includegraphics[width=\linewidth]{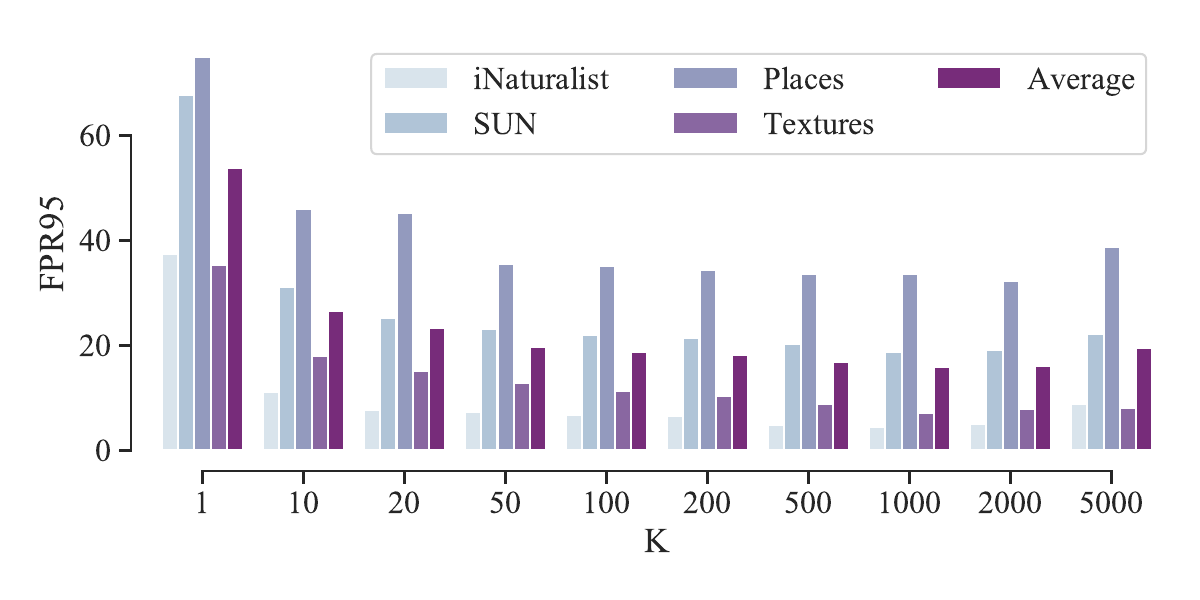}
\endminipage\hfill
\caption{\textbf{Comparison with the effect of different $k$}. (a) Small out-of-distribution detection benchmarks. (b) Large-scale out-of-distribution detection benchmarks.}\label{fig:cifar_k}\label{fig:imagenet_k}
\end{figure}

\textbf{Effect of $k$.} From our experiments, the choice of $k$ is largely dependent on the size of training dataset, which is also the size of the KNN memory bank $\mathcal{M}$. We generally suggest setting $k$ as $1$\textperthousand~of the size of training dataset. For example, the training sizes of CIFAR-10 and ImageNet are $50,000$ and $1,280,000$ respectively, and the optimal $k$ under these two settings are around $50$ and $1000$ in \figurename~\ref{fig:cifar_k} and \figurename~\ref{fig:imagenet_k} respectively.

\begin{figure}[h]
    \centering
 \minipage{0.49\linewidth}%
   \includegraphics[width=\linewidth]{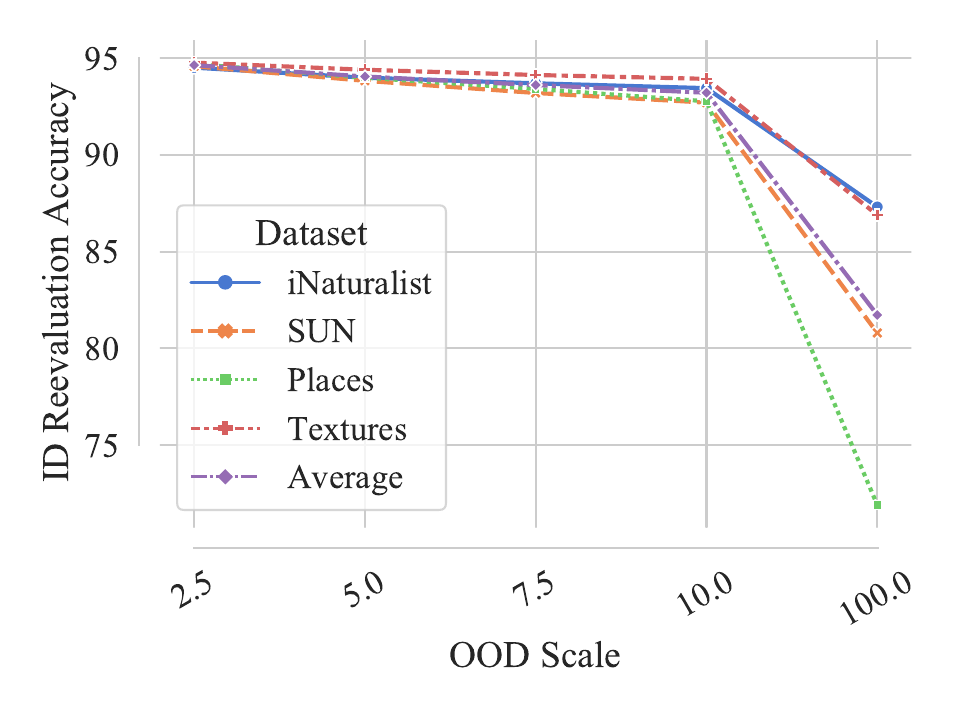}
\label{fig:imagenet_scale_in}
\endminipage\hfill
\minipage{0.49\linewidth}
 \includegraphics[width=\linewidth]{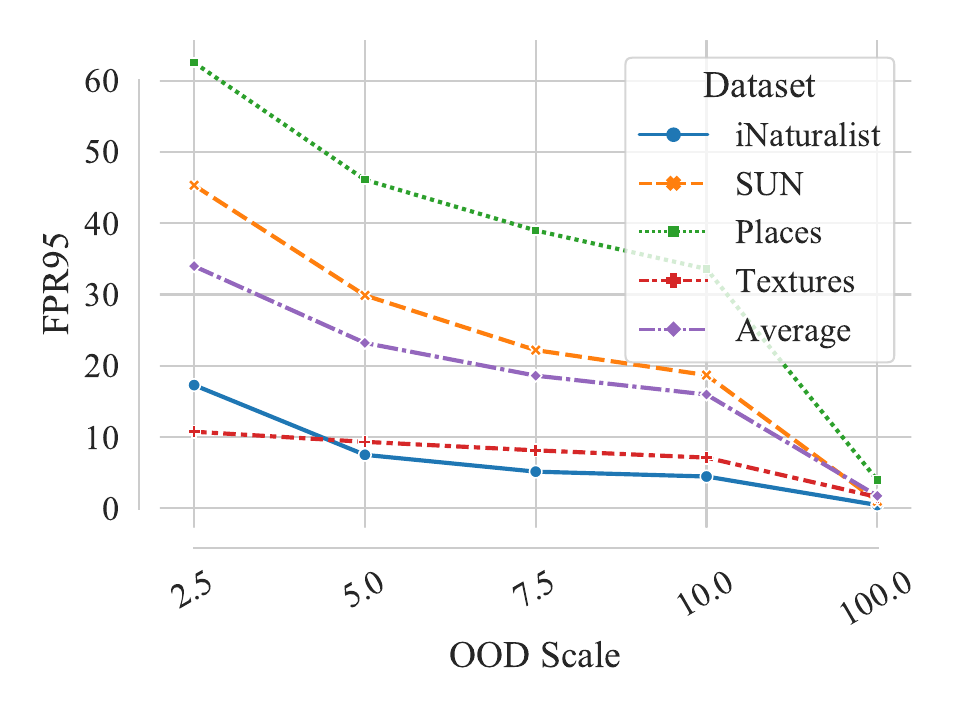}
\label{fig:imagenet_scale_out}
\endminipage\hfill
\caption{\textbf{Ablation of the effect of OOD scale $\kappa$} on (a) ID reevaluation accuracy and (b) OOD detection performance (FPR95) considering different datasets.}\label{fig:imagenet_scale}
\end{figure}

\begin{figure}[h]
    \centering
 \minipage{0.49\linewidth}%
   \includegraphics[width=\linewidth]{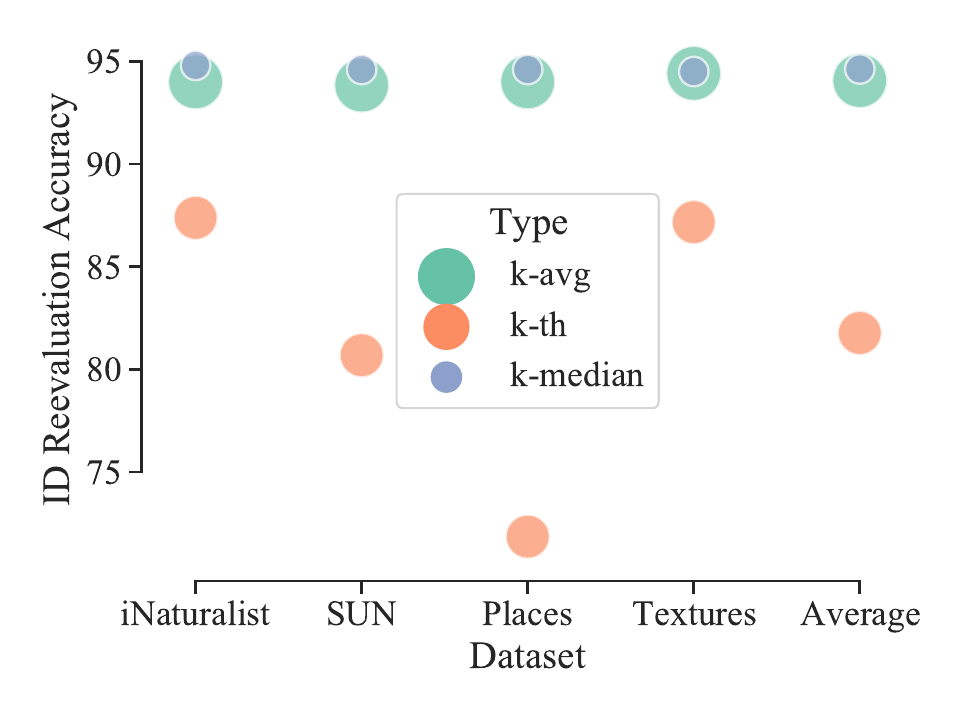}
\label{fig:imagenet_avg_in}
\endminipage\hfill
\minipage{0.49\linewidth}
 \includegraphics[width=\linewidth]{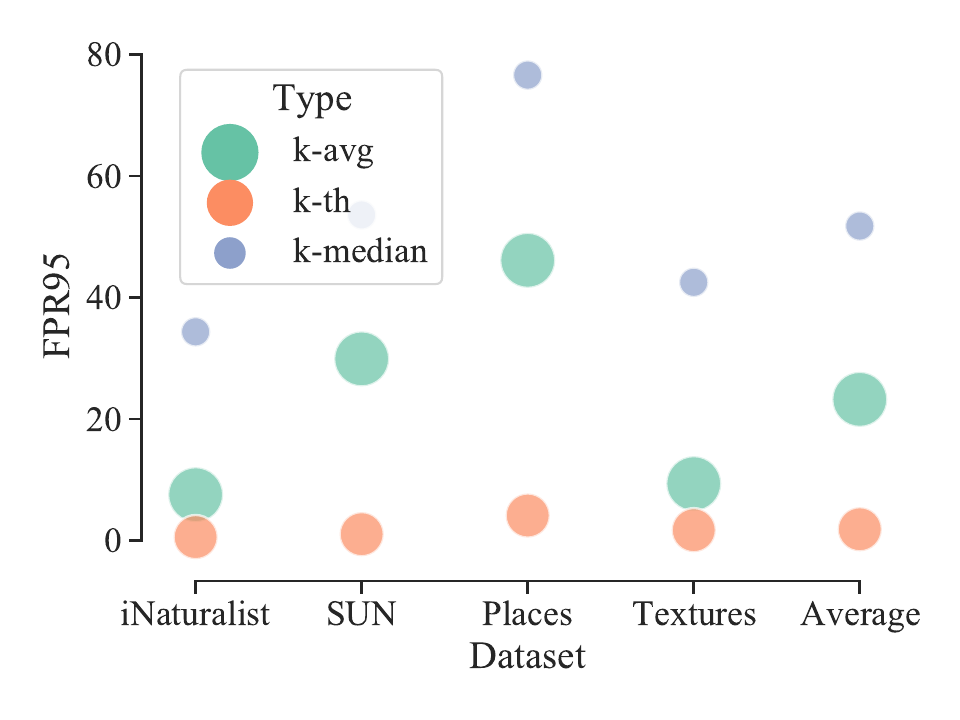}
\label{fig:imagenet_avg_out}
\endminipage\hfill
\caption{\textbf{Ablation of the effect of using $k$-avg, $k$-th, $k$-median nearest neighbor distance} considering different OOD datasets on large-scale benchmarks.}\label{fig:imagenet_avg}
\end{figure}

\textbf{Ablation of $K$ (\figurename~\ref{fig:imagenet_k}), OOD scale $\kappa$ (\figurename~\ref{fig:imagenet_scale}), OOD selection magin $\gamma$ (Table~\ref{tab:imagenet_margin}), and different methods for calculating Knn scores (\figurename~\ref{fig:imagenet_avg}).} Most phenomenons are similar to that in small scale experiments. Differently, (1) the large-scale benchmark needs a smaller margin $\gamma$ because the training dataset is large and a large margin $\gamma>1.5$ will reject all OOD samples for memory augmentation (\figurename~\ref{fig:imagenet_scale}). (2) The OOD scale $\kappa$ has a great effect on OOD performance, but has a relatively small influence on ID performance (Table~\ref{tab:imagenet_margin}). Even if we choose $\kappa=10.0$, the ID accuracy is still greater than $93\%$, and hence we choose $\kappa=10$. 
\begin{table}[]
\caption{\textbf{Ablation of the effect of selection margin $\gamma$} on ID reevaluation accuracy and OOD detection performance (FPR95) considering different  datasets.}\label{tab:imagenet_margin}
\centering
\begin{tabular}{cccccc}
\toprule
Dataset & Metric & $\gamma=1.0$ & $\gamma=1.1$ & $\gamma=1.2$ & $\gamma=\infty$ \\ \midrule
\multirow{5}{*}{\begin{tabular}[c]{@{}c@{}}ID \\ Reevaluation \\ Accuracy\end{tabular}} & iNaturalist & 93.50 & 94.20 & 94.40 & 95.00 \\
 & SUN & 92.70 & 94.20 & 94.50 & 95.00 \\
 & Places & 92.70 & 94.60 & 94.90 & 95.00 \\
 & Textures & 93.90 & 94.60 & 94.70 & 95.00 \\
 & Average & 93.10 & 94.40 & 94.65 & 95.00 \\ \midrule
\multirow{5}{*}{FPR} & iNaturalist & 4.48 & 12.28 & 22.06 & 59.08 \\
 & SUN & 18.70 & 39.17 & 48.72 & 69.53 \\
 & Places & 33.59 & 59.07 & 69.75 & 77.09 \\
 & Textures & 7.13 & 10.11 & 10.73 & 11.56 \\
 & Average & 15.97 & 30.16 & 37.81 & 54.32 \\ \bottomrule
\end{tabular}%
\end{table}

\begin{figure}[h]
    \centering
 \minipage{0.49\linewidth}%
  \includegraphics[width=\linewidth]{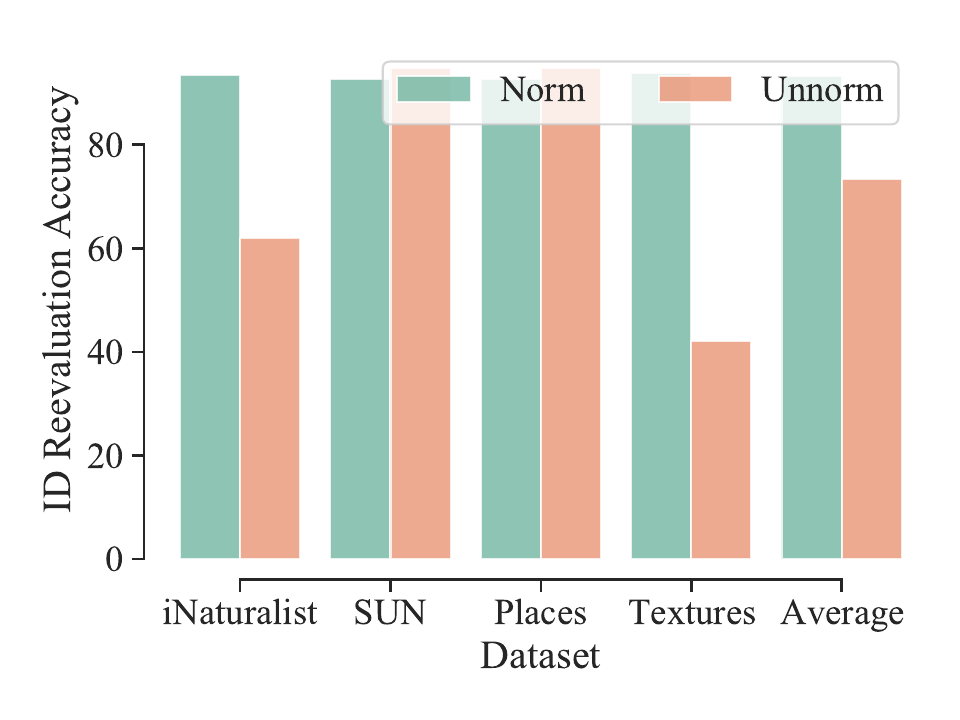}
\label{fig:imagenet_norm_in}
\endminipage\hfill
\minipage{0.49\linewidth}
 \includegraphics[width=\linewidth]{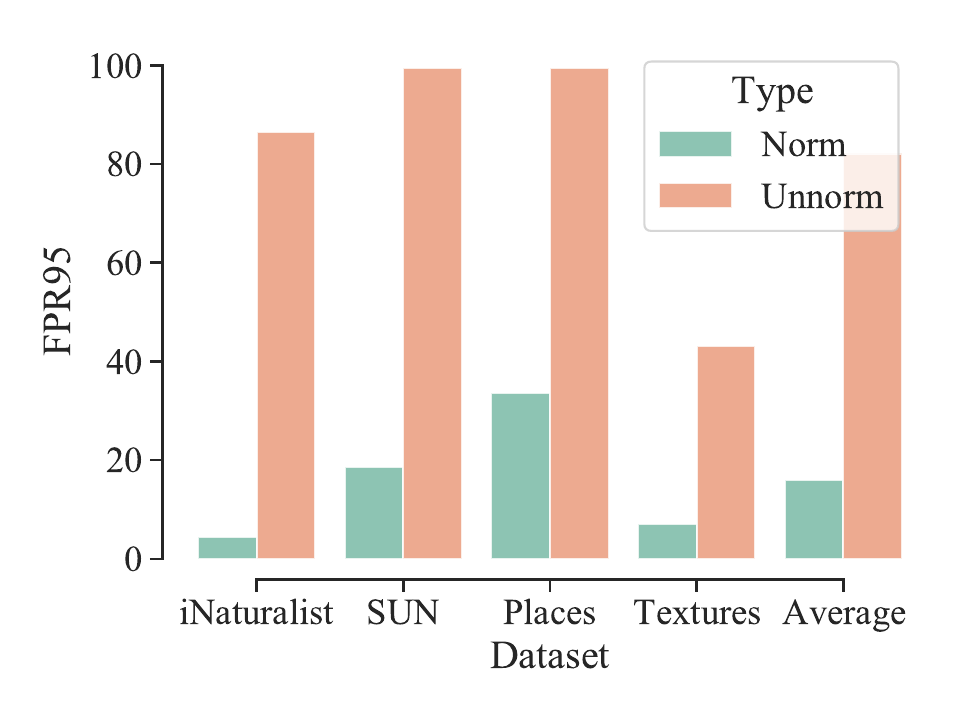}
\label{fig:imagenet_norm_out}
\endminipage\hfill
\caption{\textbf{Ablation of the effect of using normalization in the features vs. without normalization} considering different OOD datasets on large-scale benchmarks. (a) ID reevaluation accuracy. (b) FPR. }\label{fig:image_norm}
\end{figure}

\textbf{Large scale benchmark needs feature normalization before $k$-nearest neighbor search.} One major different design choice for small scale and large scale OOD detection benchmarks is the normalization process $h_\theta(x) /\parallel h_\theta(x) \parallel_2$. In the previous section, experiments show that the normalization process makes only a small difference in the FPR metric on small scale benchmarks (\figurename~\ref{fig:cifar_norm}). However, \figurename~\ref{fig:image_norm} shows that, without feature normalization, the average FPR metric on the large scale benchmark is increased by $66.2$ compared to \abbr with feature normalization. Besides, without normalization, the ID reevaluation accuracy will also be harmed. Namely, although normalization makes only small differences on some small benchmarks, not applying normalization can sometimes make the model perform very poorly.

\begin{table}[]
\caption{\textbf{Results on small-scale benchmarks} different evaluation settings. The ID dataset is cifar-10 and the metric is FPR95.} \label{tab:settings}
\centering
\resizebox{\linewidth}{!}{%
\begin{tabular}{@{}lcccc@{}}
\toprule
\textbf{Method} & \textbf{SVHN} & \textbf{LSUN} & \textbf{iSUN} & \textbf{Texture} \\ \midrule
KNN & 24.53 & 25.29 & 25.55 & 27.57 \\
AdaODD & 2.17 & 5.8 & 4.9 & 15.82 \\
AdaODD (Stream) & 2.17 & 5.37 & 2.1 & 3.83 \\
AdaODD (OODs Mixture) & 1.13 & 5.17 & 2.59 & 5.69 \\
AdaODD (ID-OOD Mixture) & 2.32 & 5.48 & 4.87 & 14.97 \\ \bottomrule
\end{tabular}%
}

\end{table}

\subsection{Evaluation on different evaluation protocols}\label{app:sec_settgns}
We have adopted an evaluation protocol that aligns with the standard practices of traditional OOD detection literature, which first calculates a threshold in a large number of ID instances and then uses it for OOD detection. We understand that there might be a concern that the model could exploit the information that all newly arrived instances are from the OOD class and simply assign OOD labels to them, leading to unfair or misleading results. 

To address this concern, we take steps to ensure a fair evaluation. In our experiments, we evaluate the \textbf{ID reevaluation accuracy}, which involves re-feeding all the ID data to the detector after the entire adaptation process. This reevaluation process is performed to validate the reliability of our approach and check for any signs of label leakage. The results presented in~\figurename~\ref{fig:cifar_magin} demonstrate that our method achieves consistent and reliable detection performance, and label leakage does not appear to be a significant concern in our setting.

Besides, the traditional evaluation protocol may be inconsistent with the actual situation during the model deployment stage. We are hence interested in the situation with OOD samples coming from several different distributions. We use three possible settings to highlight the efficiency of our proposed methods. 

\begin{itemize}
    \item \textbf{(Sequentially)} Incoming OOD samples are from different OOD distributions sequentially, namely, the algorithm is first evaluated in SVHN. After the evaluation, high-confident OOD samples in the SVHN have been stored, and we then evaluate them on LSUN. A similar process is conducted on iSUN, and Texture respectively. That is,  the incoming OOD sample originates from different auxiliary OOD distributions once.
    \item \textbf{(OODs Mixture)} The test OOD distribution is a mixture of multiple OOD distributions rather than coming from the same distribution. For the cifar-10 benchmark, we mix the SVHN, LSUN, iSUN, and Texture datasets as one OOD distribution.
    \item \textbf{(ID-OOD Mixture)} The test distribution is a mixture of one OOD distribution and the ID distribution rather than a single OOD distribution. For each dataset of the cifar-10 benchmark (e.g., SVHN, LSUN, iSUN, and Texture), we mix each of the datasets with the validation set of the ID dataset (cifar-10). In this case, the threshold $\lambda$ will be determined by the training set of cifar-10 because we need to ensure the test ID samples are not seen in advance. To ensure a fair and unbiased evaluation, we mix and shuffle all the ID and OOD data as the input stream. Note that this setting is more practical because after the detector being development, both ID and OOD samples will appear and should be detected.
\end{itemize}

Results in Table~\ref{tab:settings} show that, as the data from different OOD datasets arrive sequentially, the performance of our algorithm improves with increasing OOD samples, especially on the last evaluated dataset Texture, where a significant improvement is achieved. When all OOD datasets are mixed together, the final results of all OOD datasets are significantly improved. That is to say, as long as there exists an auxiliary OOD distribution, our algorithm can utilize it to improve the detection performance on other OOD samples. With a mixture of ID and OOD samples, the performance of \abbr will not be harmed because we only augment the memory bank with highly confident ID samples, which further enhance the quality of the K-nearest neighbor search.

\begin{table}[]
\caption{\textbf{FPR95 of different covariate settings.}}\label{tab:covariate}
\resizebox{\columnwidth}{!}{%
\begin{tabular}{cccccc}
\toprule
 &  & \textbf{SVHN} & \textbf{LSUN} & \textbf{iSUN} & \textbf{Texture} \\ \midrule
\multicolumn{2}{c}{\abbr} & 2.17 & 5.8 & 4.9 & 15.82 \\
\multirow{2}{*}{Gaussian noise} & severity=1 & 2.142 & 5.617 & 4.972 & 15.627 \\
 & severity=5 & 2.173 & 5.632 & 4.834 & 15.782 \\ \toprule
\end{tabular}%
}
\end{table}

\subsection{Whether covariance shifts affect the performance of \abbr?}
Because there is currently no existing benchmark specifically designed for testing the impact of covariance shift in the ID samples during test time. To address this, we conduct experiments using the cifac-10-c dataset to simulate the scenario you mentioned, where there is a covariance shift in the ID samples during test time.

In our experiments, we mix the Gaussian noise part of the cifac-10-c dataset with various OOD datasets and evaluate its effect on OOD sample detection. The results, as shown in the Table.~\ref{tab:covariate}, indicate that the impact of covariate shift on OOD detection is relatively small, as represented by different severity values that indicate varying degrees of covariate shift. This finding is understandable because, compared to covariate shift, label shift can cause more significant differences in feature representations, making it easier to distinguish the OOD samples.


In summary, our experiments suggest that the impact of covariate shift on OOD detection is relatively minor compared to other factors like label shift, which can result in more pronounced differences in feature representations and, consequently, a higher potential for detection. Nonetheless, further research and benchmarks dedicated to investigating covariate shift's influence on OOD detection will be valuable for a more comprehensive understanding of this topic.

\section{Conclusion}



This paper underscores the limitations of conventional threshold-based adjustments when handling OOD data without prior knowledge. Relying solely on adjustments based on ID data often falls short of achieving the optimal decision boundary. However, our research demonstrates that leveraging test-time samples offers a promising avenue to enhance detector refinement and ultimately improves overall detection performance. This observation is further corroborated by recent, somewhat frustrating theoretical findings~\cite{fang2022is}, which have motivated us to explore a new paradigm: test-time adaptation for OOD detection.

While existing TTA methods exist, they are not directly applicable to the domain of OOD detection. Our attempts to adapt current OOD detectors into TTA versions yield limited performance gains. As a solution, we introduce a novel approach for OOD detection, referred to as \abbr. This pioneering method harnesses online target samples for OOD detection and employs a K-Nearest Neighbors (KNN) classifier with memory augmentation, providing robust and effective OOD detection capabilities. Both theoretical analyses and empirical experiments underscore the efficacy of our proposed approach. Across various benchmarks, including small- and large-scale datasets, our method consistently outperforms several strong baselines and evaluation settings in terms of detection performance. This work opens new avenues for enhancing OOD detection, emphasizing the importance of test-time adaptation in this context.

{\small
\bibliographystyle{IEEEtran}
\bibliography{main}
}





\end{document}